\documentclass[10pt,journal,compsoc]{IEEEtran}
\ifCLASSOPTIONcompsoc
  \usepackage[nocompress]{cite}
\else
  \usepackage{cite}
\fi
\ifCLASSINFOpdf
  \usepackage[pdftex]{graphicx}
\else
\fi
\usepackage{url}

\usepackage{hyperref} 
\hypersetup{
  colorlinks=true,
  linkcolor=magenta,
  urlcolor=blue,
  citecolor=blue,
}

\usepackage{enumitem}

\usepackage[acronym]{glossaries}
\makeglossaries
\newacronym{hsr}{HSR}{High Spatial Resolution}

\usepackage{pifont}

\usepackage{fontawesome}
\usepackage{float}
\usepackage{amsmath,amsfonts}
\usepackage{algorithmic}
\usepackage{algorithm}
\usepackage{listings}
\usepackage{array}
\usepackage{subfigure}
\usepackage{textcomp}
\usepackage{stfloats}
\usepackage{url}
\usepackage{verbatim}
\usepackage{graphicx}
\usepackage{cite}
\usepackage{xcolor}
\usepackage{multirow}
\usepackage{booktabs}
\usepackage{makecell}
\usepackage{threeparttable}
\usepackage[left,mathlines]{lineno}

\definecolor{generalcolor}{HTML}{412F8A}
\definecolor{rscolor}{HTML}{fc8e62}
\definecolor{ques}{RGB}{192,0,0}
\definecolor{revision}{RGB}{0,0,0}
\newcommand{\generalcolor}[1]{\textcolor{generalcolor}{#1}}
\newcommand{\rscolor}[1]{\textcolor{rscolor}{#1}}
\newcommand{\gb}{\generalcolor{$\bullet$\,}}
\newcommand{\rsb}{\rscolor{$\bullet$\,}}

\begin{document}
%
\title{EarthVL: A Progressive Earth Vision-Language Understanding and Generation Framework}
%
%
%
%

\author{Junjue Wang,~\IEEEmembership{Graduate Student Member,~IEEE,}
Yanfei Zhong,~\IEEEmembership{Senior Member,~IEEE,}
Zihang Chen,
Zhuo Zheng,
Ailong Ma,~\IEEEmembership{Member,~IEEE,}
and Liangpei Zhang,~\IEEEmembership{Fellow,~IEEE}
\thanks{Corresponding author: Yanfei Zhong. E-mail: zhongyanfei@whu.edu.cn}
\thanks{The preliminary versions of this work were presented in NeurIPS 2021 D\&B Track \cite{wang2021loveda} and AAAI 2024\cite{wang2024earthvqa}}
}

%
%

\markboth{Journal of \LaTeX\ Class Files,~Vol.~14, No.~8, August~2015}%
{Shell \MakeLowercase{\textit{et al.}}: Bare Demo of IEEEtran.cls for Computer Society Journals}
%



\IEEEtitleabstractindextext{%
\begin{abstract}
Earth vision, as a cutting-edge research topic in artificial intelligence, has achieved many milestones in geospatial object recognition. However, there has been a lack of sufficient exploration of object-relational reasoning, limiting the ability to understand remote sensing scenes comprehensively.
To address this, a progressive Earth vision-language understanding and generation framework is proposed, including a multi-task 
dataset (EarthVLSet) and a semantic-guided network (EarthVLNet).
Focusing on city planning applications, 
EarthVLSet includes 10.9k sub-meter resolution remote sensing images, land-cover masks, and 
761.5k textual pairs involving both multiple-choice and open-ended visual question answering (VQA) tasks.
In an object-centric way, EarthVLNet is proposed to progressively achieve 
semantic segmentation, relational reasoning, and comprehensive understanding.
The first stage involves land-cover segmentation to generate object semantics for VQA guidance.
Guided by pixel-wise semantics, the object awareness based large language model (LLM) performs 
relational reasoning and knowledge summarization to 
generate the required answers.
As for optimization, the numerical difference loss is proposed to dynamically add difference penalties, addressing the various
objects' statistics.
Three benchmarks including semantic segmentation, multiple-choice, and open-ended VQA demonstrated the superiorities of EarthVLNet,
yielding three future directions: 1)
segmentation features consistently enhance VQA performance even in cross-dataset scenarios; 2) multiple-choice tasks show greater sensitivity to the vision encoder than to the language decoder; and 3) open-ended tasks necessitate advanced vision encoders and language decoders for an optimal performance.
We believe this dataset and method will provide a beneficial benchmark that connects ``image-mask-text'', advancing geographical applications for Earth vision. 
Data and code are available \href{https://github.com/Junjue-Wang/EarthVL}{here}.

\end{abstract}
}

\maketitle

\IEEEdisplaynontitleabstractindextext

%
\IEEEpeerreviewmaketitle

\IEEEraisesectionheading{\section{Introduction}\label{sec:introduction}}

%
%
%
%

\begin{figure}[!hbt]
  \centering
  \includegraphics[width=1.0\linewidth]{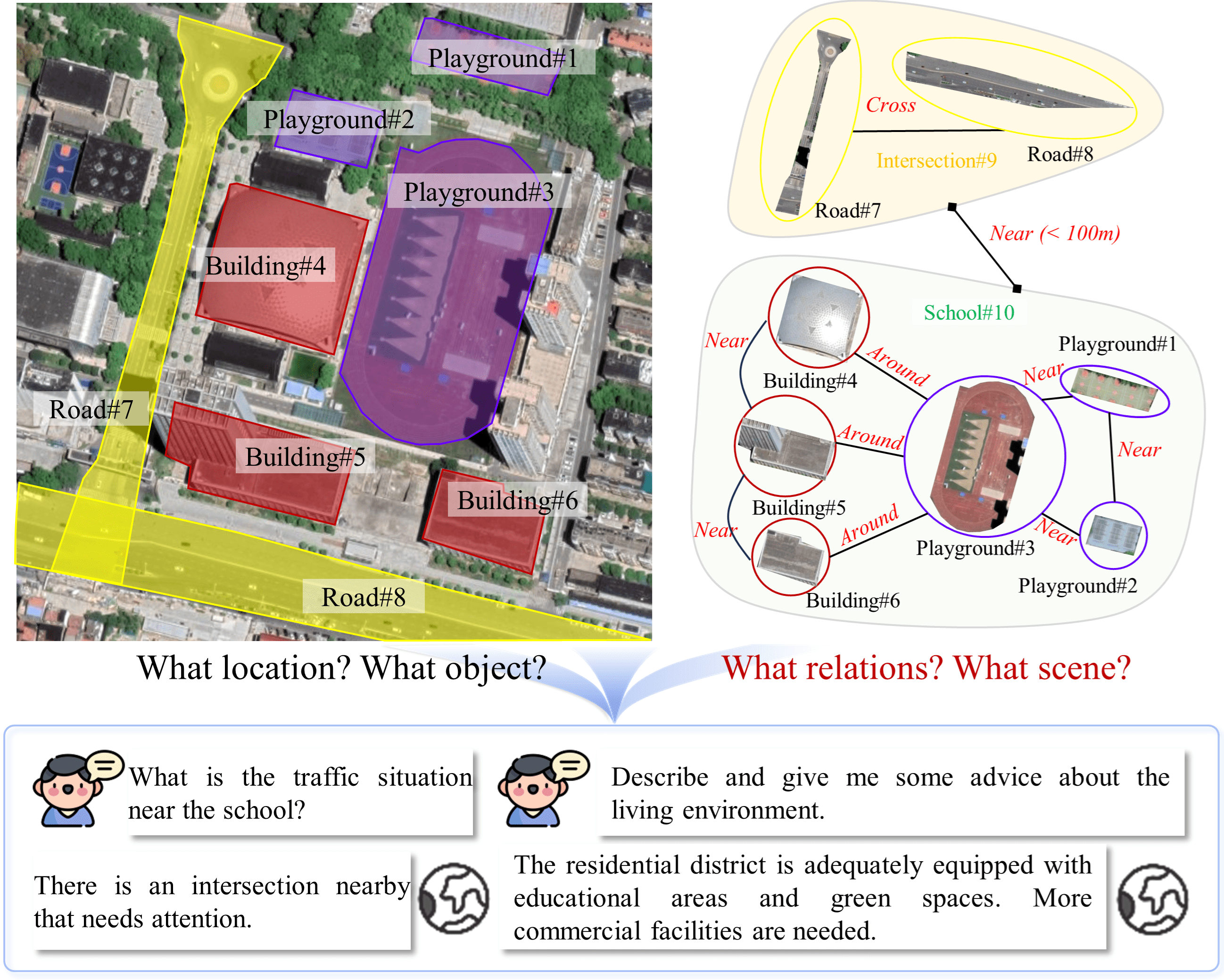}
  \vspace{-0.5cm}
  \caption{Comprehensive understanding of HSR remote sensing imagery.
  To automatically achieve ``what locations have what objects'' and ``what relations form what scenes'', we propose a benchmark dataset and method to connect the semantic segmentation and VQA tasks.
  }
  \label{fig:4w-goal}
\vspace{-0.1cm}
\end{figure}

\IEEEPARstart{H}{igh} spatial resolution (HSR) Earth observation platforms continuously provide massive remote sensing images, 
displaying the geometries, details, and textures of geospatial objects clearly. 
Earth vision focuses on developing artificial intelligence algorithms to assist 
humans in interpreting large-scale HSR images and involves
many fields, including scene classification \cite{xia2017aid}, aerial object detection \cite{ding2021object}, and land-cover semantic segmentation\cite{wang2021loveda}.
Scene classification is aimed at learning the global land-use types, 
and detection as well as segmentation obtains the categories and locations of the local objects.
However, most tasks ignore the spatial and semantic relations between objects and struggle with comprehensive reports\cite{gao2022cric}.
Leveraging the powerful reasoning capabilities of large language models (LLMs),
we aim to comprehend HSR images holistically, 
enabling progressive and interactive assistance in city planning.
As illustrated in Fig.~\ref{fig:4w-goal},
HSR image understanding can be divided into two key aspects, i.e.,
``\textit{what locations have what objects?}'' and ``\textit{what relations form what scenes?}''.
To address these questions, we first employ semantic segmentation algorithms 
to accurately extract the geospatial object locations and categories, generating pixel-level semantic results.
Building upon the object semantics,
we then introduce visual question answering (VQA)\cite{10146482} methods that enable LLM-based relational reasoning and textual generation. 
By performing these tasks simultaneously, decision-makers can gain a holistic understanding of HSR scenes from \textit{both intuitive visual and linguistic aspects}.

Focusing on city planning requirements,
our questions are designed to align 
with UN-Habitat's Sustainable Development Goals (SDG)\cite{sgd2020} assessment tools, involving housing, climate, traffic, water system, etc.
By analyzing urban landscapes, infrastructures, and spatial patterns based on remote sensing images, city planners can make effcient decisions for urban development
\cite{cai2025linking}.
Detecting urban villages in developed countries provides an effective solution for 
governing illegal buildings and neglected urban spaces for cultivating vegetables\cite{wang2026cityvlm}.
Analysis of green space within residential areas contributes to improving vegetation distributions
 and alleviates urban heatwaves\cite{yin2023unequal}.
To this end, we integrate the city planning tasks that remote sensing images could facilitate and
propose a progressive Earth vision-language understanding and generation framework.

The contributions of this paper are listed as follows:
\begin{enumerate}
\item \textbf{Multi-Task EarthVLSet.} 
A multi-task vision-language dataset (\textit{EarthVLSet}) has been curated, covering 17 countries on six continents worldwide.
The \textit{EarthVLSet} includes 10.9k HSR images, land-cover semantic masks, and 
734k question-answer (QA) pairs with multiple-choice and open-ended VQA tasks embedded.
The multiple-choice questions include 
eight types, ranging from easy basic judging to complex relational reasoning, and even more challenging comprehensive analysis. 
The open-ended VQA questions require 
city planning and decision-making answers with varied lengths. 
\textit{EarthVLSet} connects the ``\textit{image-mask-QA pairs}''
to facilitate effective Earth vision understanding.

\item \textbf{Semantic-Guided EarthVLNet.}
\textit{EarthVLNet} progressively learns the representations of land-cover semantic segmentation and VQA. 
The land-cover segmentation network is first trained to provide semantic guidance.
By leveraging pixel-level semantics, the object awareness based LLM can reason out the refined spatial and semantic relations, 
significantly improving the VQA performance on complex types. 
Compared to the traditional cross-entropy loss, the object counting enhanced optimization introduces the numerical difference sensitivity,
addressing the various objects' statistics in HSR scenes.

\textit{EarthVLNet} unifies multiple-choice and open-ended VQA tasks within a single framework, significantly enhancing model flexibility and applicability to real-world scenarios.

\item \textbf{Benchmarks and Insights.}
Based on \textit{EarthVLSet}, three HSR remote sensing benchmarks have been established systematically, involving 
semantic segmentation (18 methods), multiple-choice VQA (16 methods), and open-ended VQA (8 methods) tasks.
Our comprehensive analysis yields three significant insights:
1) segmentation features demonstrate general applicability to VQA tasks, maintaining their utility even in cross-dataset scenarios;
2) multiple-choice VQA tasks benefit predominantly from powerful vision encoders, while exhibiting less sensitivity to the complexity of language decoders;
and 3) open-ended VQA tasks necessitate both robust vision encoders and advanced language decoders for an optimal performance.

\end{enumerate}

Preliminary versions of this work were published in \cite{wang2021loveda} and \cite{wang2024earthvqa}.
We have extended the dataset and method in terms of several aspects.
Firstly, we have expanded the data scope from the original three cities in China to a global scale, covering 17 countries worldwide.
Secondly, more types of QA pairs have been added, evaluating
the model's ability to perceive the spatial layouts and directions of geospatial objects.
The open-ended QA pairs have been designed to promote complex summarization.
Thirdly, we have developed semantic-guided 
LLMs to achieve sophisticated relational reasoning and variable-length generation.
Fourthly, the object counting tasks have been separately modeled to avoid training conflicts.
Last but not least, 
the systematic benchmark results also reveal several promising directions for future improvements.

\begin{figure*}[hbt]
\centering
\subfigure[Global visual features]{
\includegraphics[width=0.30\linewidth]{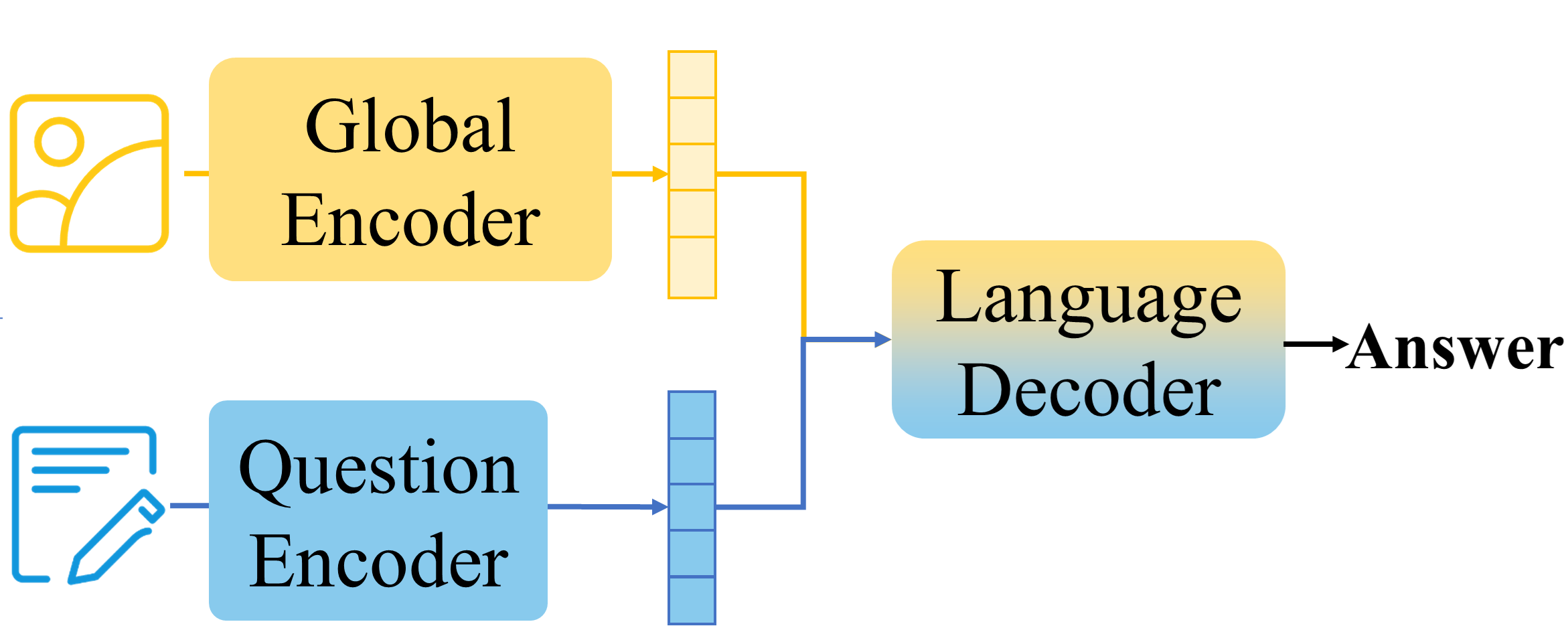}
\label{fig:related_work:sub1}
}
\subfigure[Bounding box features]{
\includegraphics[width=0.32\linewidth]{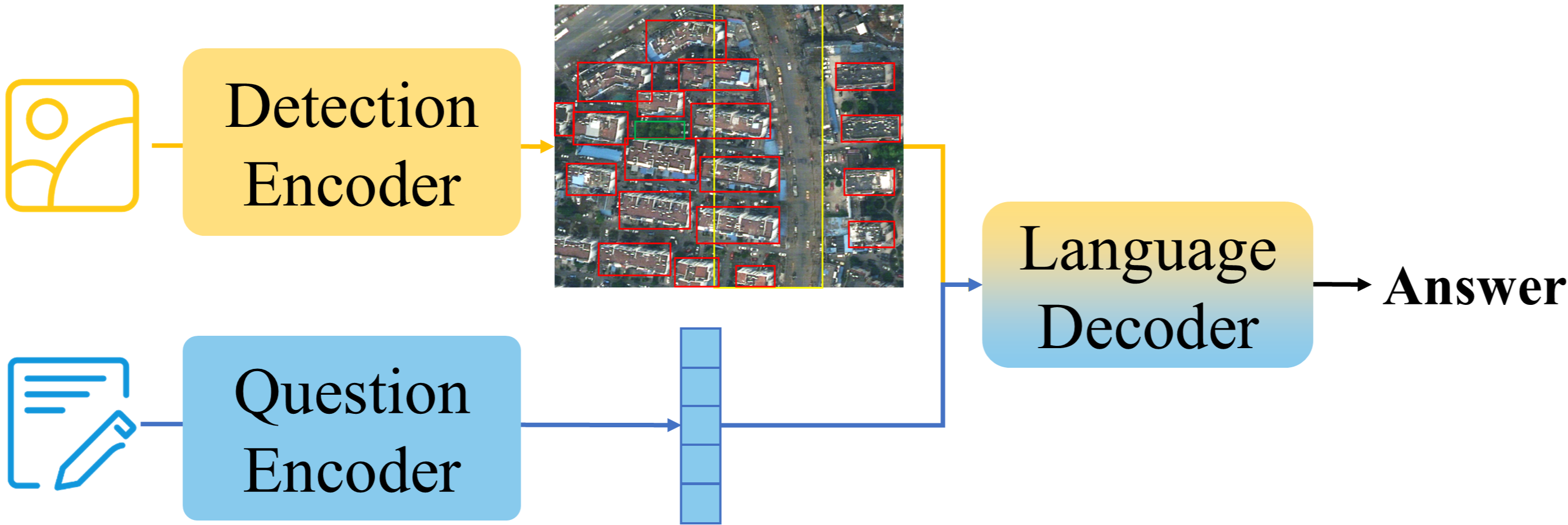}
\label{fig:related_work:sub2}
}
\subfigure[Semantic segmentation features (Proposed)]{
\includegraphics[width=0.32\linewidth]{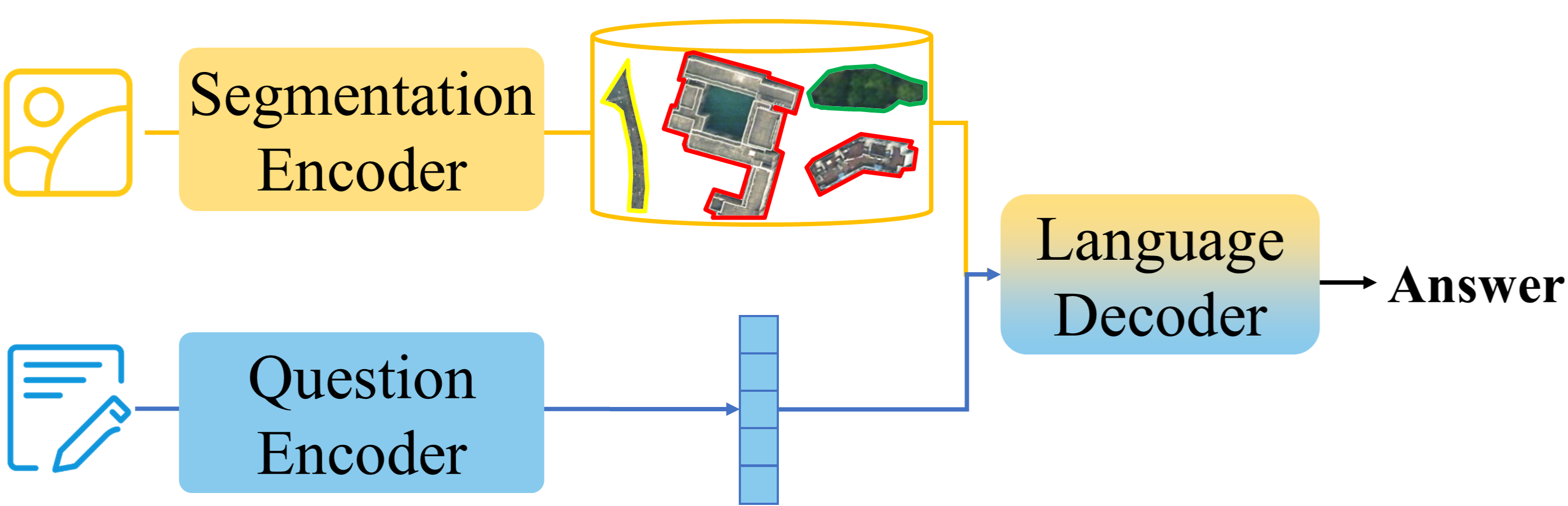}
\label{fig:related_work:sub3}
}
\vspace{-0.2cm}
\caption{\textcolor{revision}{The VQA methods can be divided into three categories, according to the vision feature type: (a) global fusion methods, (b) bounding box based methods, and (c) segmentation-based methods. 
The segmentation features provide more refined semantic boundaries at the pixel level, contributing accurate object statistics and relational reasoning for complex HSR scenes.
}}
\vspace{-0.3cm}
\label{fig:related_work}
\end{figure*}

\begin{table*}[!hbt]
  \centering
  \resizebox{1.0\linewidth}{!}{
  \begin{threeparttable}
  \caption{Comparison Between EarthVLSet and the Existing Remote Sensing VQA Datasets.} \label{tab:cpm_data}
  \begin{tabular}{lllll|c|c|cccccccc}
  \hline
  Dataset   & Image size   & Res.(m)    & \#QAs   & \#Images & OE & SM  & BJ & BC & CJ & CC & AE & DisA & DirA & CA \\ \hline
  RSVQA-LR \cite{lobry2020rsvqa}  & 256    & 10    & 77K    & 772 & $\times$  & $\times$ & $\checkmark$  & $\checkmark$  & $\checkmark$  & $\checkmark$  & $\times$ & $\times$   & $\times$  & $\times$ \\
  RSVQA-HR \cite{lobry2020rsvqa}   & 512    & 0.15    & 955K   & 10659  & $\times$  & $\times$ & $\checkmark$  & $\checkmark$  & $\checkmark$  & $\checkmark$  & $\times$ & $\times$   & $\times$  & $\times$  \\
  RSVQAxBen \cite{lobry2021rsvqa}  & 120    & 10--60   & 15M    & 590326 & $\times$  & $\times$ & $\checkmark$  & $\times$  & $\checkmark$  & $\checkmark$  & $\times$ & $\times$   & $\times$  & $\times$  \\
  RSIVQA \cite{zheng2021mutual}   & 512--4000  & 0.3--8   & 111K   & 37000 & $\times$ & $\times$ & $\checkmark$  & $\checkmark$  & $\checkmark$  & $\times$  & $\checkmark$ & $\times$   & $\times$  & $\times$  \\
  HRVQA \cite{li2023hrvqa}    & 1024     & 0.08    & 1070K  & 53512  & $\times$ & $\times$  & $\checkmark$  & $\times$  & $\checkmark$  & $\checkmark$ & $\times$   & $\times$  & $\times$  & $\times$  \\
  TextRS-VQA \cite{bashmal2023visual}    & 256     & 0.06--5    & 6245  & 2143  & $\times$ & $\times$  & $\checkmark$  & $\checkmark$  & $\checkmark$  & $\checkmark$ & $\times$   & $\times$  & $\times$  & $\times$  \\
  CDVQA \cite{yuan2022change}    & 512    & 0.5--3   & 122K   & 2968 & $\times$  & $\checkmark$ & $\checkmark$  & $\checkmark$  & $\times$  & $\times$  & $\times$ & $\times$   & $\times$  & $\times$  \\
  FloodNet \cite{rahnemoonfar2021floodnet}  & 3000--4000 & -     & 11K   & 2343 & $\times$  & $\checkmark$ & $\times$  & $\checkmark$  & $\checkmark$  & $\checkmark$  & $\checkmark$ & $\times$   & $\times$  & $\times$ \\
  RescueNet-VQA \cite{RESCUENet-VQA} & 3000--4000 & 0.15    & 103K   & 4375  & $\times$  & $\checkmark$ & $\times$  & $\checkmark$  & $\checkmark$  & $\checkmark$  & $\checkmark$ & $\times$   & $\times$  & $\times$ \\ \hline
  EarthVQA \cite{wang2024earthvqa}   & 1024     & 0.3     & 208K   & 6000 & $\times$  & $\checkmark$ & $\checkmark$  & $\checkmark$  & $\checkmark$  & $\checkmark$  & $\checkmark$ & $\times$   & $\times$  & $\checkmark$  \\
  EarthVLSet & 1024     & 0.3     & 761K   & 10950  & $\checkmark$ & $\checkmark$ & $\checkmark$  & $\checkmark$  & $\checkmark$  & $\checkmark$  & $\checkmark$ & $\checkmark$   & $\checkmark$  & $\checkmark$  \\ \hline
  \end{tabular} 
  \begin{tablenotes}
    \footnotesize 
    \item Abbreviations:  OE -- open-ended, SM -- semantic mask, BJ -- basic judging, BC -- basic counting, CJ -- complex judging, CC -- complex counting, AE -- attribute extraction, DisA -- distribution analysis, DirA -- directional analysis, CA -- comprehensive analysis.
  \end{tablenotes}
  \end{threeparttable}}
  \vspace{-0.3cm}
\end{table*}

\vspace{-0.2cm}
\section{Related Work}
\subsection{Land-Cover Semantic Segmentation}
In the context of deep learning,
fully convolutional networks (FCNs)
have dominated the HSR land-cover mapping fields.
Considering the multi-scale objects,
ResUNet\cite{diakogiannis2020resunet} incorporates residual connections, atrous convolutions, and pyramid scene parsing pooling to capture contextual features.
LinkNet\cite{chaurasia2017linknet} and UNet++\cite{zhou2018unet++} further improve the multi-scale extraction capabilities by adding 
more cross-level connections.
Semantic-FPN\cite{SFPN} employs a feature pyramid structure and asymmetric decoder to
fuse the multi-scale features effectively.
By reformulating the encoder-decoder structure,
HRNet\cite{wang2020deep} implements a multi-scale design into every layer. 
To suppress background false alarms,
FarSeg\cite{farsegpp} and FactSeg\cite{FactSeg} introduce additional object-scene relations to activate 
the objects of interest, enhancing the representation of small objects.
To capture long distance dependencies,
UNetFormer\cite{wang2022unetformer} combines a Swin-Transformer and ResNet for efficient urban mapping.
In this paper, we report the benchmark results of 16 CNN and Transformer segmentation methods, 
providing robust semantic features for VQA.
Equipped with pixel-wise guidance, remote sensing VQA can further explore the 
intricate relations between objects through human interaction.

\vspace{-0.2cm}
\subsection{Visual Question Answering}
\indent\textbf{Variant Visual Features.}
VQA methods can be divided into three categories, based on the visual feature type (Fig.~\ref{fig:related_work}).
\textit{(a) Global fusion methods.}
Early research considered VQA as the fusion of global visual and language features \cite{antol2015vqa}.
The visual and language features are individually processed by a CNN and a Recurrent Neural Network, 
and the global features are fused by a language decoder to predict the final answer.
The stacked attention network (SAN) \cite{yang2016stacked} and the memory, attention, and composition network (MAC) \cite{hudson2018compositional}
are the typical structures.
\textit{(b) Bounding box based methods.}
To reason refined relations efficiently, bottom-up and top-down (BUTD)\cite{Anderson_2018_CVPR} 
uses Faster-RCNN features to incorporate object features.
The bounding boxes serve as restricted mechanisms, enabling the fusion model to effectively capture key objects in the scene.
The modular co-attention network (MCAN) \cite{yu2019mcan} employs Transformers for vision-language feature interaction, while 
the learning cross-modality encoder representations from transformers
(LXMERT) framework \cite{tan2019lxmert} uses a triplet encoder to explore the intra- and cross-modality relations. 
D-VQA \cite{wen2021debiased} addresses textual bias through a unimodal bias detection module. 
Other approaches \cite{gao2022transform} incorporate external knowledge bases to enhance the generalizability. 
In addition, VQA applications have expanded from single-frame images to video interactions \cite{gao2021env}.
In conclusion, the global fusion methods overlook the local object semantics, and the
bounding box based approaches inevitably include irrelevant background details, especially for irregular objects such as roads, rivers, and forests.
To address these issues,
our \textit{(c) segmentation-based method} utilizes pixel-level semantics with more precise boundaries and richer details.

\noindent \textbf{Vision-Language Generation Model.}
Early approaches treated the VQA task as multi-choice classification, predicting answers based on maximum output probabilities \cite{10445007}. 
However, these approaches lack flexibility and struggle with complex scenarios such as scene descriptions or urban planning advice.
To produce variable-length responses, 
the open-ended VQA methods replace the classifier with a generative model such as 
the long short-term memory
(LSTM) \cite{LSTM} or Transformer \cite{vaswani2017attention}.
ViLBERT \cite{lu2019vilbert} uses dual BERTs for vision and language processing, followed by co-attention Transformer layers for cross-modal interaction. 
ViLT \cite{kim2021vilt} streamlines this process by using a unified Transformer for both modality interaction and answer generation.
Equipped with outstanding reasoning abilities,
LLMs have showcased superior performances in answer generation, deriving 
many instruction-tuning vision-language models (VLMs), e.g., 
Flamingo \cite{alayrac2022flamingo}, BLIP-2 \cite{BLIP2}, InstructBLIP \cite{instructblip}, LLaVA \cite{liu2024visual}, and LLaVANeXT \cite{liu2024llavanext}.
By utilizing the pre-trained VLMs, 
the injected learnable parameters can be fine-tuned on VQA datasets, effectively adapting the conditional generation tasks.
Because VLMs can achieve variable-length responses,
we adopt these  models to obtain extensive city planning advice.
By introducing semantic guidance and numerical optimization, 
\textit{EarthVLNet} can address the intricate relations between various geospatial objects.  

\vspace{-0.3cm}
\subsection{Visual Question Answering in Remote Sensing}
The remote sensing community has made significant strides in VQA, developing various datasets and methods.
As for datasets, the RSVQA \cite{lobry2020rsvqa} dataset includes remote sensing images and the georeferenced Open Street Map (OSM) properties.
By designing QA templates, answers can be automatically generated by querying the OSM fields.
Guided by the 2018 CORINE Land Cover product \cite{dimitrov2019satellite}, the RSVQAxBen dataset \cite{lobry2021rsvqa} was constructed by judging and area estimation.
By compiling the existing HSR detection and classification datasets, the
RSIVQA dataset \cite{zheng2021mutual} automatically generates QA pairs from their semantic annotations.
To increase the diversity, the TextRS-VQA dataset \cite{bashmal2023visual} is made up of images from classification datasets (AID \cite{xia2017aid}, PatternNet \cite{zhou2018patternnet} and NWPU-RESISC45 \cite{cheng2017remote}) with manually annotated QA pairs.
The CDVQA dataset \cite{yuan2022change} introduces a bi-temporal change detection VQA task.
Constructed from the SECOND dataset \cite{yang2021asymmetric}, the semantic changes are queried automatically from the bi-temporal masks.
Focusing on disaster assessment, the FloodNet \cite{rahnemoonfar2021floodnet} and RescueNet-VQA \cite{RESCUENet-VQA} datasets
provide QA pairs for the damage to roads and buildings.
Methodologically, many approaches have adapted general VQA techniques to remote sensing. 
RSIVQA \cite{zheng2021mutual} introduces mutual attention for improved multi-modal interaction. 
To promote open-world tasks,
SenCLIP \cite{jain2024senclip} and GRAFT \cite{mallremote} integrate remote sensing and grounding images with open-world textual prompts, 
achieving land-use mapping. 
SkyScript \cite{wang2024skyscript} adopts the OSM database to introduce open-world semantics for multi-object recognition. 
Recent open-ended advancements include RSGPT \cite{hu2023rsgpt}
which fine-tunes the projector of InstructBLIP, and GeoChat \cite{kuckreja2023geochat}, 
which applies low-rank adaptation (LoRA) \cite{hu2021lora} to fine-tune LLaVA on multi-task datasets, 
creating unified VLMs for remote sensing applications.

The existing remote sensing vision-language research typically 
focuses on simple conversations.
Compared to the existing datasets shown in Tab.~\ref{tab:cpm_data}, 
\textit{EarthVLSet} offers three key advantages:
\textbf{1) Multi-level annotations.} These involve pixel-level semantic labels, object-level reasoning questions, and scene-level analysis. This multi-perspective supervision assists with comprehensive representation.
\textbf{2) Complex and practical questions.} While the existing datasets mainly focus on counting and judging questions involving simple relational reasoning about one or two object types, \textit{EarthVLSet} introduces complex analysis by incorporating spatial or semantic reasoning of more than three object types. 
The complex questions (e.g., distances, layouts, topologies, sub-properties) are designed to meet the needs of city planning.
\textbf{3) Open-ended VQA.} \textit{EarthVLSet} includes open-ended VQA labels, training models to generate indefinite-length answers. 
This facilitates the summarization of comprehensive descriptions and renovation advice.


\begin{figure*}[!hbt]
  \centering
  \includegraphics[width=1.0\linewidth]{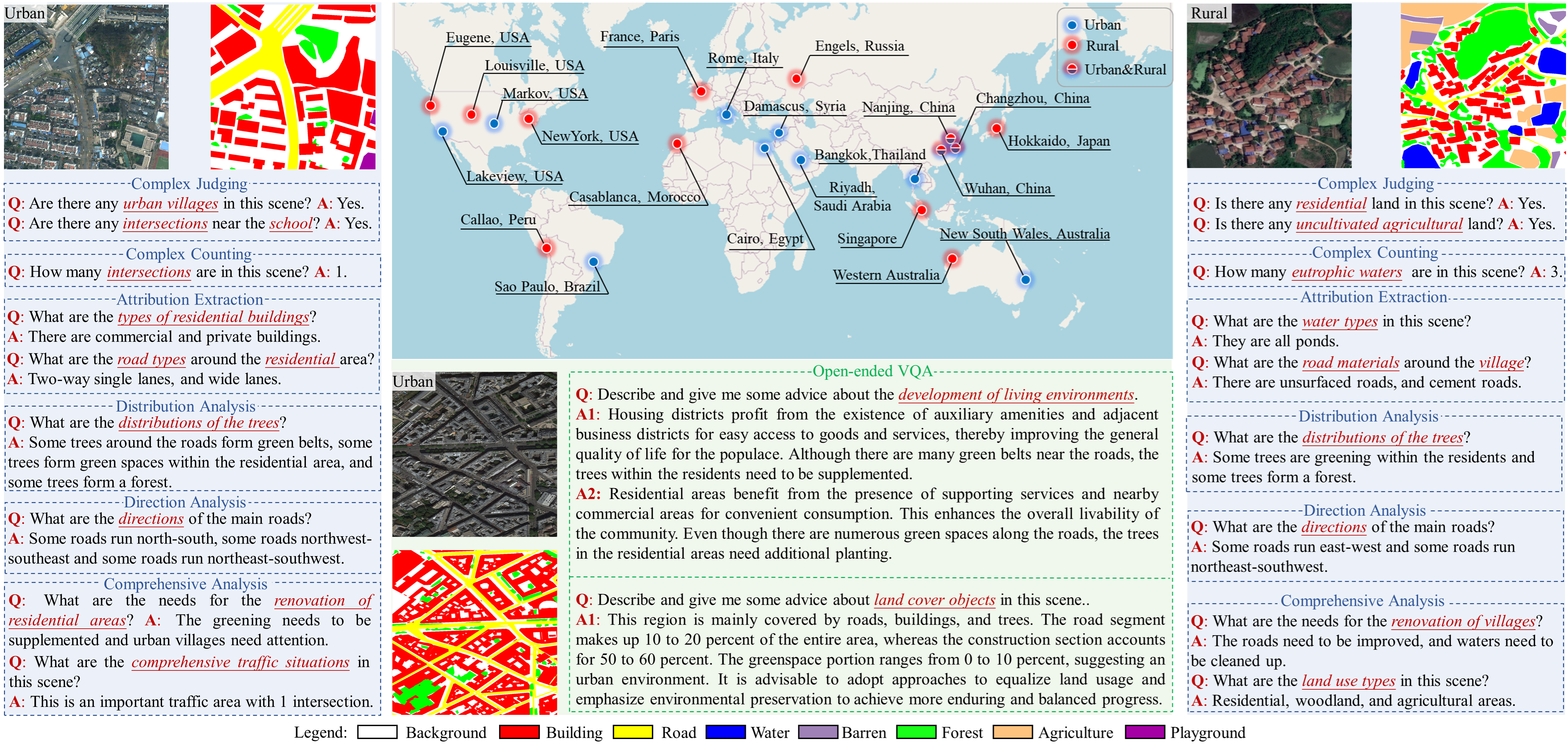}
  \vspace{-0.7cm}
  \caption{Global distribution of the city planning-oriented EarthVLSet dataset. The different regions represent diverse object landscapes, spectra, and affordances, challenging the model transferability. 
  The multi-choice QA pairs require relational reasoning (topologies, distances, sub-properties, etc.) for geospatial objects. The open-ended QA pairs provide detailed sentences for scene understanding from different aspects. 
  As an intermediary, the semantic mask links the remote sensing imagery and geographical language
  }
  \vspace{-0.3cm}
  \label{fig:global_dis}
\end{figure*}
\section{Multi-Task EarthVLSet}

In the following,
we detail the statistics and annotation procedures of \textit{EarthVLSet} for
multiple-choice (\S \ref{sec:3:multiple}) and open-ended VQA (\S \ref{sec:3:open}) data.

  
\begin{figure*}[!hbt]
  \centering
  \includegraphics[width=0.8\linewidth]{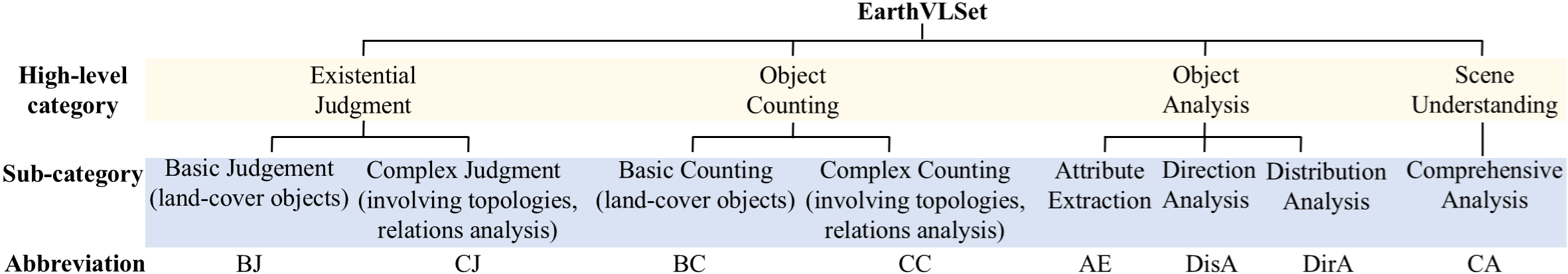}
  \vspace{-0.3cm}
  \caption{Hierarchical structures of multiple-choice question categories in EarthVLSet.
  }
  \label{fig:hierarchical_struct}
  \vspace{-0.3cm}
\end{figure*}
\vspace{-0.2cm}
\subsection{Multiple-Choice VQA Data}
\label{sec:3:multiple}
\textbf{Semantic Masks.}

Following the \textit{LoveDA} dataset, we selected eight common land-cover types for annotation, i.e., building, road, water, forest, agriculture, barren, playground, and background. The professional remote sensing annotators were trained to follow the guidelines: 1) All clearly visible objects in the seven categories (except background) must be annotated using polygons; 2) Each polygon must match the object's visual boundary; 3) Adjust image zoom as needed for precise boundary annotation; 4) Report unclear/difficult objects to team supervisors for discussion and consensus; 5) All work should be done using ArcGIS geospatial software.

For the 19 extended areas out of China, each single-area land-cover annotation required approximately 26 hours, totaling 494 person-hours. The annotation process included multiple quality checks: first, self-examination and cross-examination to correct false labels, missing objects, and inaccurate boundaries. Team supervisors then performed a quality inspection on 800 randomly sampled images, with unqualified annotations undergoing refinement. Finally, several statistics (e.g. object numbers per image, object areas, etc.) were computed to double-check the outliers. Based on DeepLabV3, preliminary experiments were conducted to ensure the validity of the annotations.
Compared to previous version,
\textit{EarthVLSet} expands the coverage from 566.231 $\rm km^2$ to 2434.793 $\rm km^2$, increasing annotated pixels by $\approx$ 1.84 times.
Because we followed the original setting to collect urban and rural images in equal proportions,
the classes show similar distributions in the different datasets.
\begin{figure}[!hbt]
  \centering
  \includegraphics[width=1.0\linewidth]{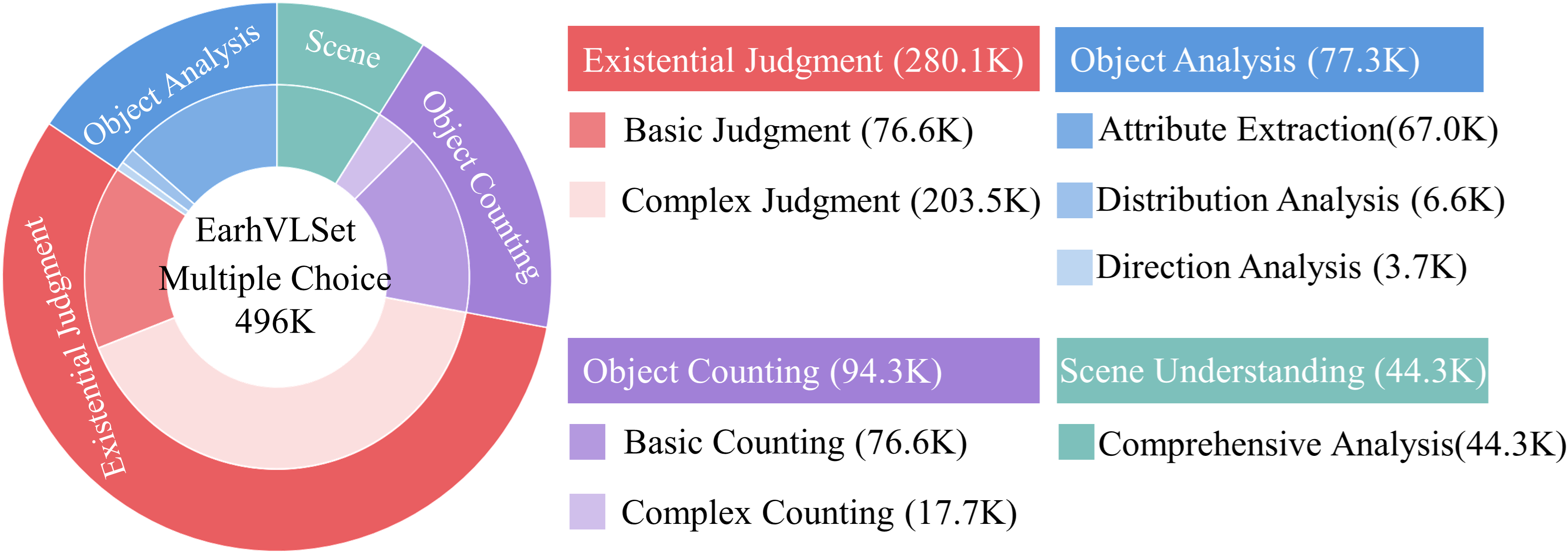}
  \vspace{-0.8cm}
  \caption{Distributions of multiple-choice questions in EarthVLSet dataset.}
  \label{fig:vqa-ques-distribution}
  \vspace{-0.8cm}
\end{figure}

\noindent \textbf{Question Distributions.}

For intuitive, we construct a hierarchical structure (Fig.~\ref{fig:hierarchical_struct}) to organize our multiple-choice questions based
on task properties and question difficulties.
Fig.~\ref{fig:vqa-ques-distribution} shows the hierarchical distributions of question samples for detailed statistics.
The existential judgment questions are designed to judge the ``existing or not'' of objects, where
\textit{basic judgment} questions only estimate the basic objects with basic land-cover types in \textit{LoveDA} dataset, and 
\textit{complex judgment} questions involve spatial and semantic reasoning between several objects.
The object counting questions also follows this `basic/complex' principal.
As for complex questions, ``Are there any intersections in this scene?'' requires topology analysis between the different roads, and 
``How many irregular buildings are in this scene?'' needs geometric analysis of the buildings.
These two questions both require spatial reasoning of ground objects.
As for semantic reasoning, ``Are there any eutrophic waters in this scene?'' requires sub-property recognition of the water bodies.
The diverse spatial and semantic reasoning requirements in the complex questions promote the model representation from different aspects.
The object analysis questions focus on the situations of key components in city planning.
The \textit{attribute extraction} questions focus on the sub-property recognition, as some key objects represent different situations  
in different geographic environments.
The \textit{distribution analysis} and \textit{direction analysis} questions aim to evaluate 
the model capability for positional awareness.
Specifically, ``What are the directions of the main roads?'' requires the model to recognize and gather all the directions in the road segments.
The \textit{comprehensive analysis} questions 
involve relational reasoning with more than two types of objects, requiring complex 
traffic evaluation, urban renovation, agricultural irrigation analysis, etc.
Due to the diverse questions with different complexities,
\textit{EarthVLSet} can measure multiple perspectives of VQA models.

\begin{figure*}[!hbt]
  \centering
  \subfigure[{\scriptsize Basic Judgment}]{
  \includegraphics[width=0.17\linewidth]{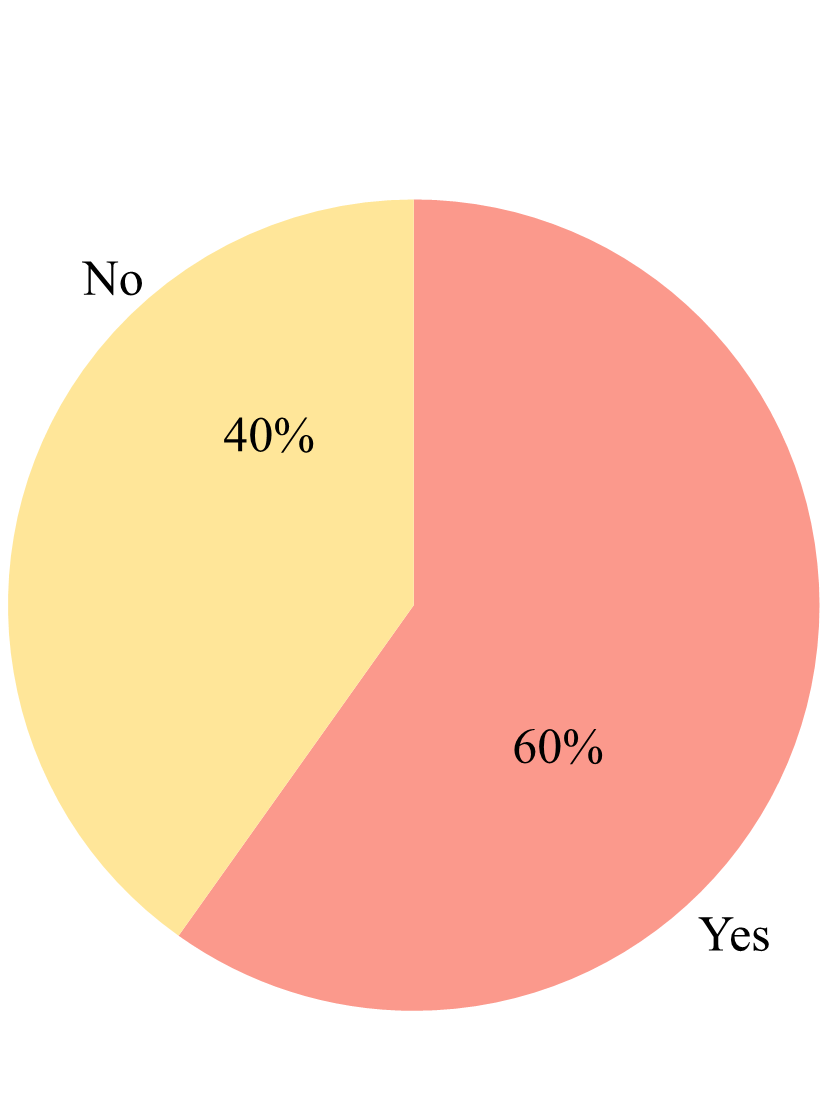}
  \label{fig:mc_ans:sub1}
  }
  \subfigure[{\scriptsize Complex Judgment}]{
  \includegraphics[width=0.17\linewidth]{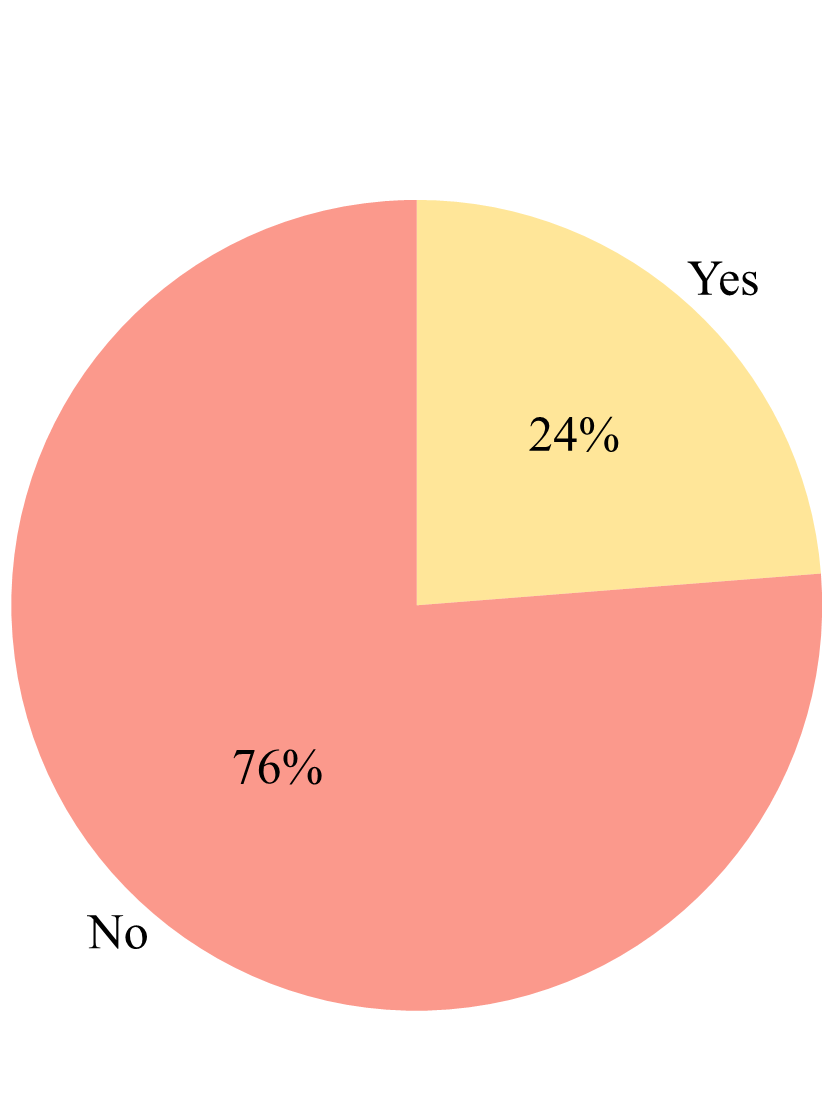}
  \label{fig:mc_ans:sub2}
  }
  \subfigure[{\scriptsize Basic Counting}]{
  \includegraphics[width=0.26\linewidth]{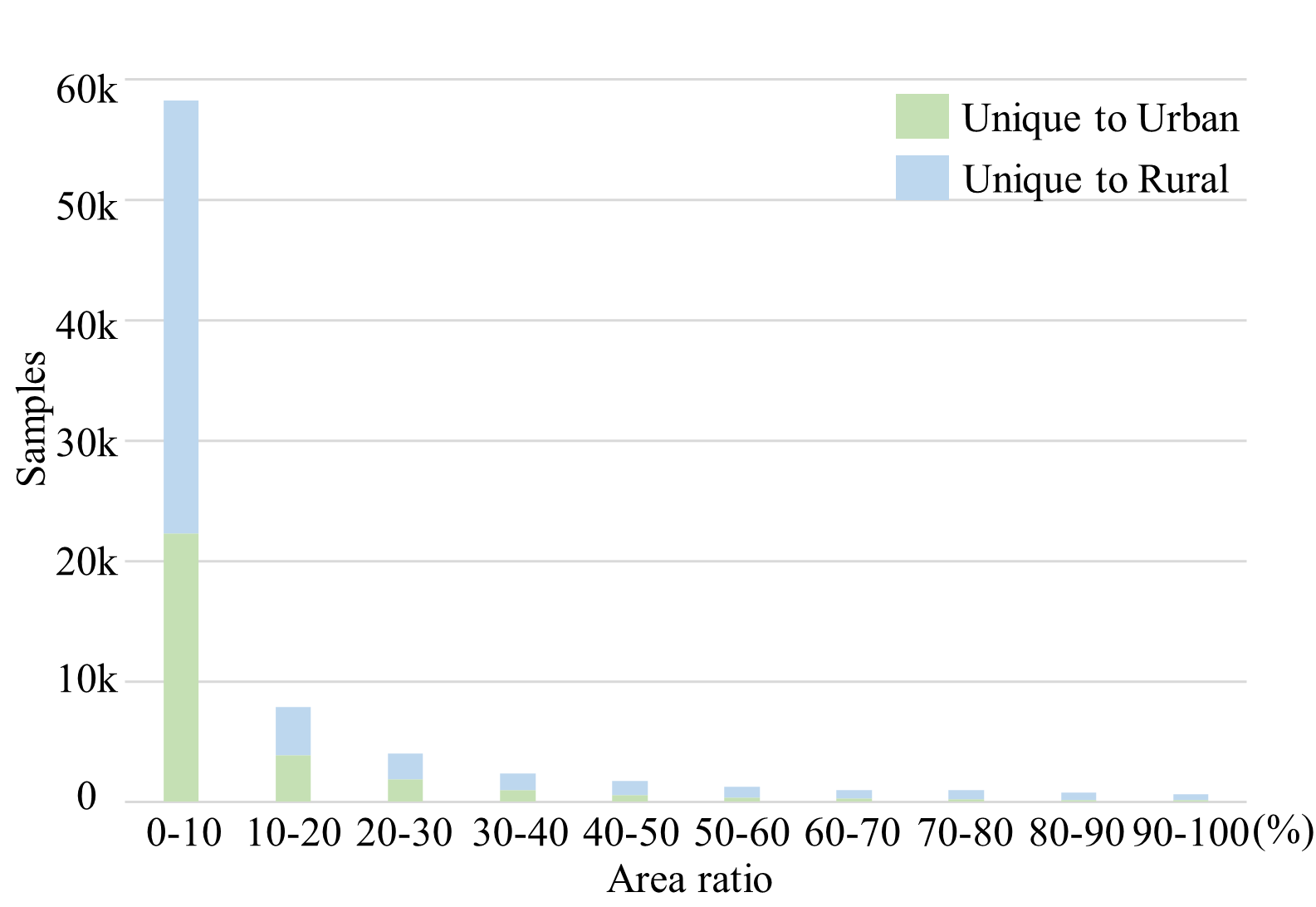}
  \label{fig:mc_ans:sub3}
  }
  \subfigure[{\scriptsize Object Situation Analysis}]{
  \includegraphics[width=0.32\linewidth]{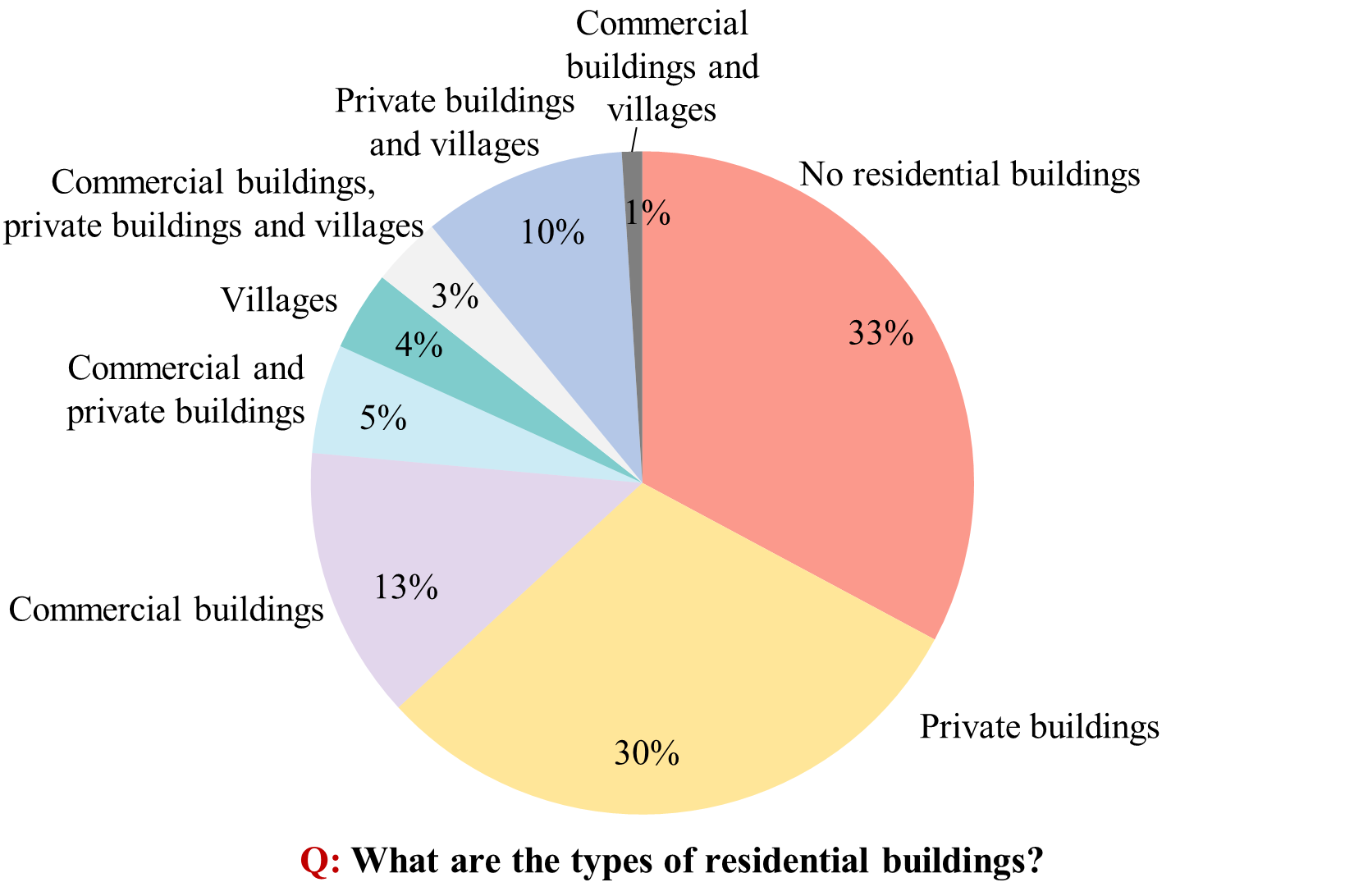}
  \label{fig:mc_ans:sub4}
  }
  \subfigure[{\scriptsize Distribution Analysis}]{
  \includegraphics[width=0.32\linewidth]{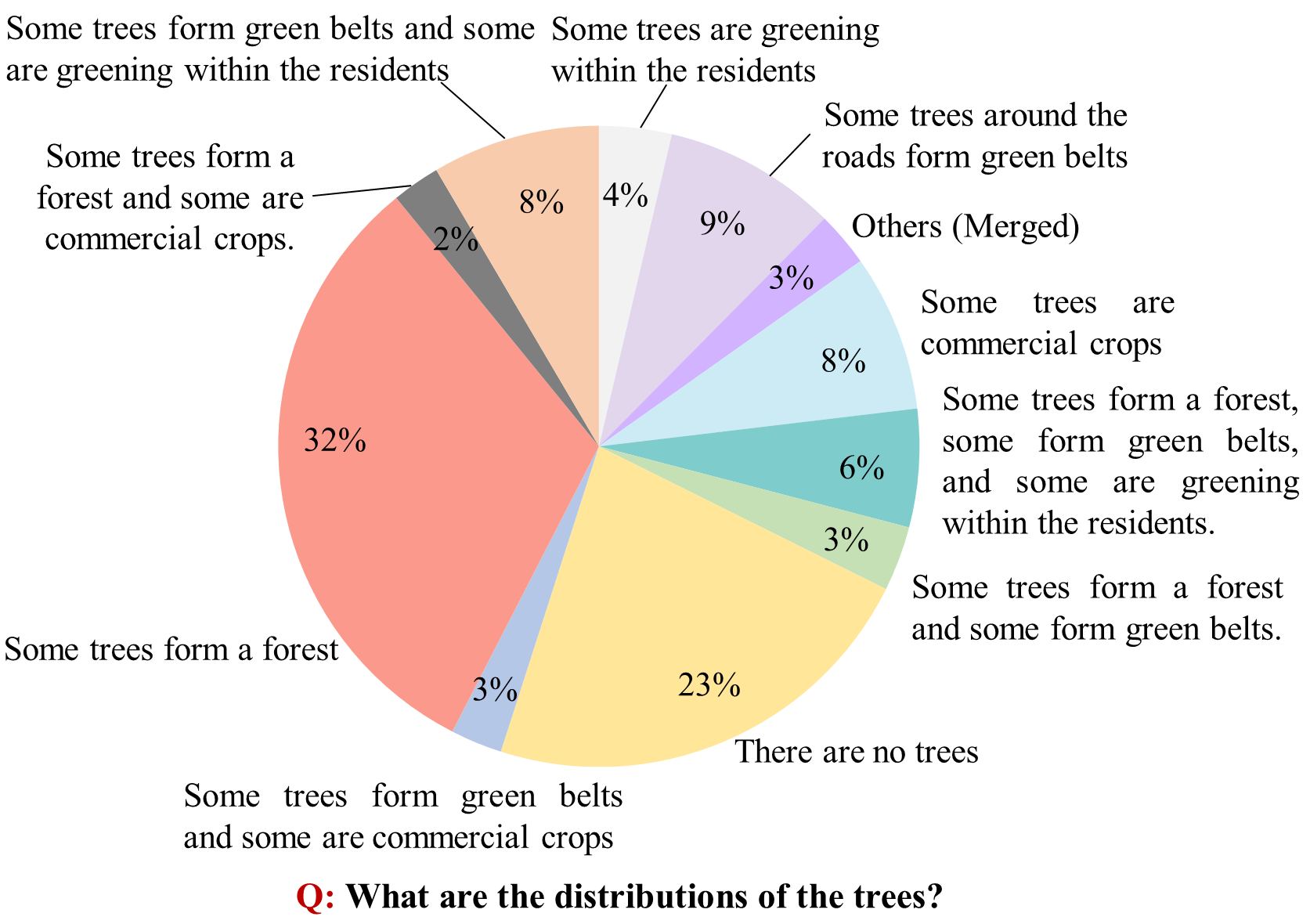}
  \label{fig:mc_ans:sub5}
  }
  \subfigure[{\scriptsize Direction Analysis}]{
  \includegraphics[width=0.30\linewidth]{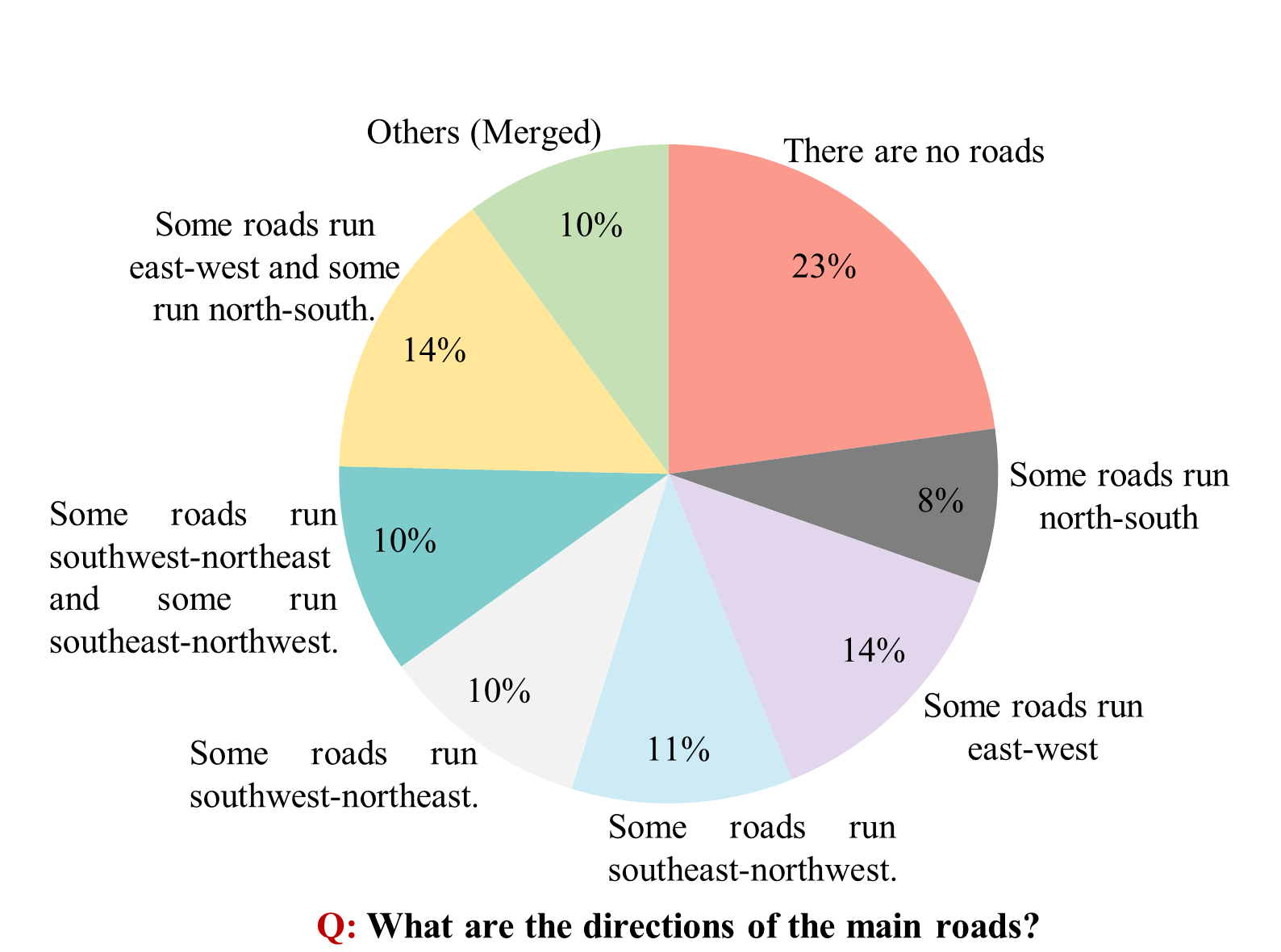}
  \label{fig:mc_ans:sub6}
  }
  \subfigure[{\scriptsize Comprehensive Analysis}]{
  \includegraphics[width=0.31\linewidth]{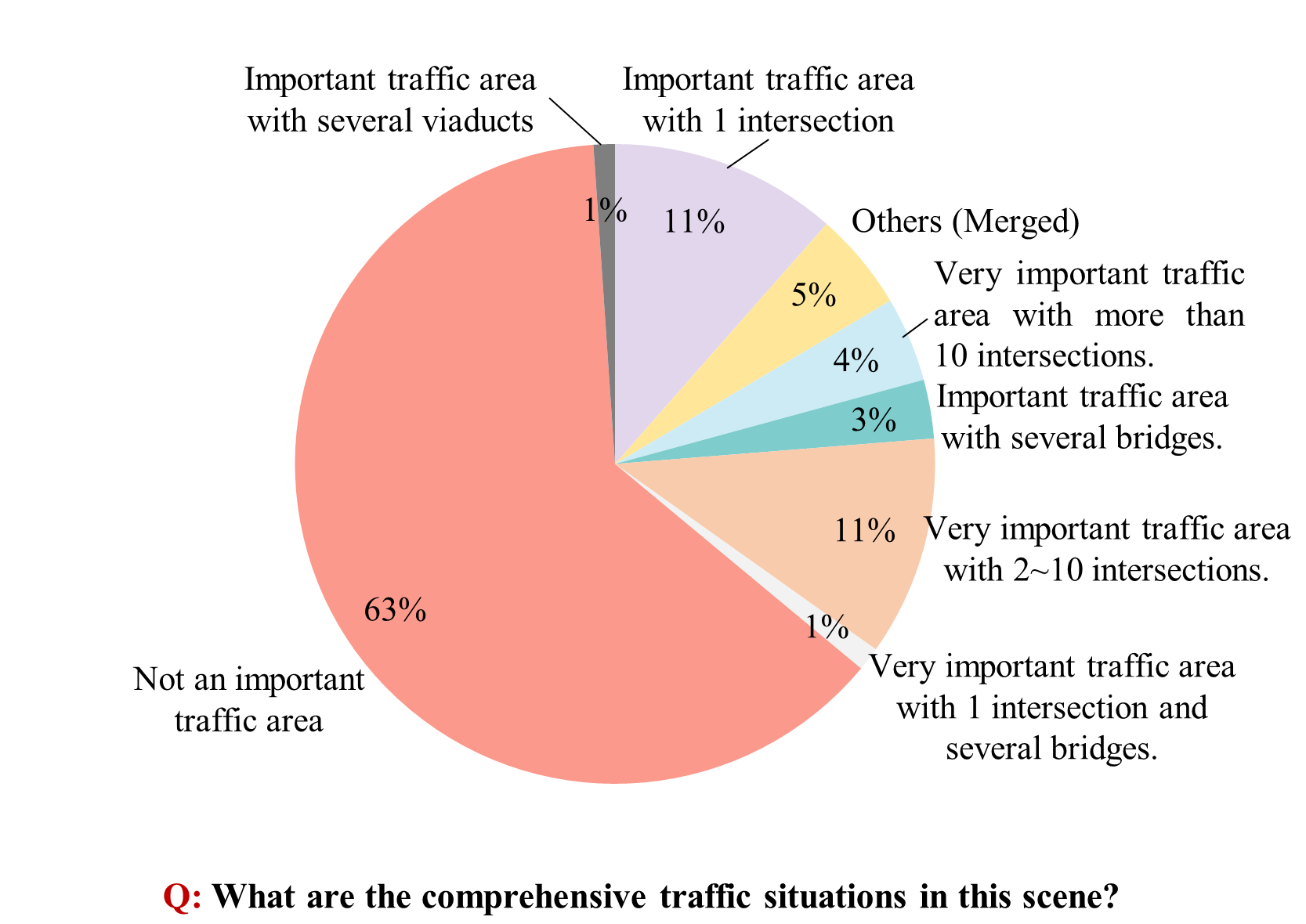}
  \label{fig:mc_ans:sub7}
  }
  \caption{Representative distributions of the multiple-choice answers with different types. 
  For a better visualization,
  some over-length answers are simplified and unusual answers are merged into ``Others''.
  The multiple-choice questions include both balanced and imbalanced answer distributions, reflecting new challenges in actual Earth environments.
  }
  \label{fig:mc_ans}
  \vspace{-0.3cm}
\end{figure*}

\noindent \textbf{Answer Distributions.}
As for the answer statistics, the representative distributions with the different types are shown in Fig.~\ref{fig:mc_ans}.
In the \textit{basic judging} answers, affirmative responses (``Yes'') constitute a majority (60\%), 
while for the complex judgment questions, negative answers (``No'') predominate (76\%).
The \textit{basic counting} answers exhibit a pronounced imbalance, characterized by a long-tail distribution.
This phenomenon indicates the spatial characteristics of HSR scenes, where objects of interest typically occupy limited areas and are dispersed across the image, 
reflecting the complexity of environments containing multiple small objects. 
Notably, the distribution patterns in urban and rural scenes demonstrate similar trends.
Regarding the \textit{object situation analysis}, the answer distribution for the question "What are the types of residential buildings?" is presented in Fig.~\ref{fig:mc_ans:sub4}. 
Private dwellings show a higher prevalence than commercial structures, which is a finding attributable to the higher population density and, consequently, the smaller spatial footprint of commercial edifices. 
The presence of private buildings and villages can also be observed in various urban contexts, including urban villages and peripheral areas.
Fig.~\ref{fig:mc_ans:sub5} depicts the distribution analysis answers to "What are the distributions of the trees?". 
Forests, being predominant in rural landscapes, account for the largest proportion of answers (32\%). 
It is noteworthy that road green belts (9\%), economic trees (8\%), and residential greening (8\%) exhibit 
comparable proportions, second only to forests, representing common arboreal distributions in both urban and rural scenes.
As shown in Fig.~\ref{fig:mc_ans:sub6},
the answer distributions of the road direction analysis are relatively balanced.
Fig.~\ref{fig:mc_ans:sub7} illustrates the answer distributions for the comprehensive analysis question, i.e., "What are the comprehensive traffic situations in this scene?". Regarding critical traffic infrastructure, 
the data indicate a higher prevalence of intersections, compared to bridges and viaducts. 
In conclusion, the multiple-choice questions include both balanced and imbalanced answer distributions, 
reflecting new challenges in actual Earth environments.

\noindent \textbf{Annotation Guidelines and Quality Control.}
According to the data division, all the images were allocated to professionally trained annotators. To ensure quality control, 
we implemented a comprehensive evaluation pipeline following the methodology outlined in \cite{wang2025disasterm3,xuan2025dynamicvl}.
Annotators were tasked with responding to all the assigned questions based on our predefined template and guidelines. Following the initial labeling phase, multiple rounds of inspection were conducted, including self-assessment, peer review, and random spot checks by team leaders. 
All samples underwent multiple revisions until they met the requisite quality standards.
For the basic judging and counting questions, answers were derived directly from the semantic masks via an automated programmatic pipeline. The annotation process for a single image required $\approx$ 10 minutes to answer all the questions.
Finally, we computed the statistics for the questions and answers to double-check the outliers. 

\begin{figure}[!hbt]
  \centering
  \includegraphics[width=1.0\linewidth]{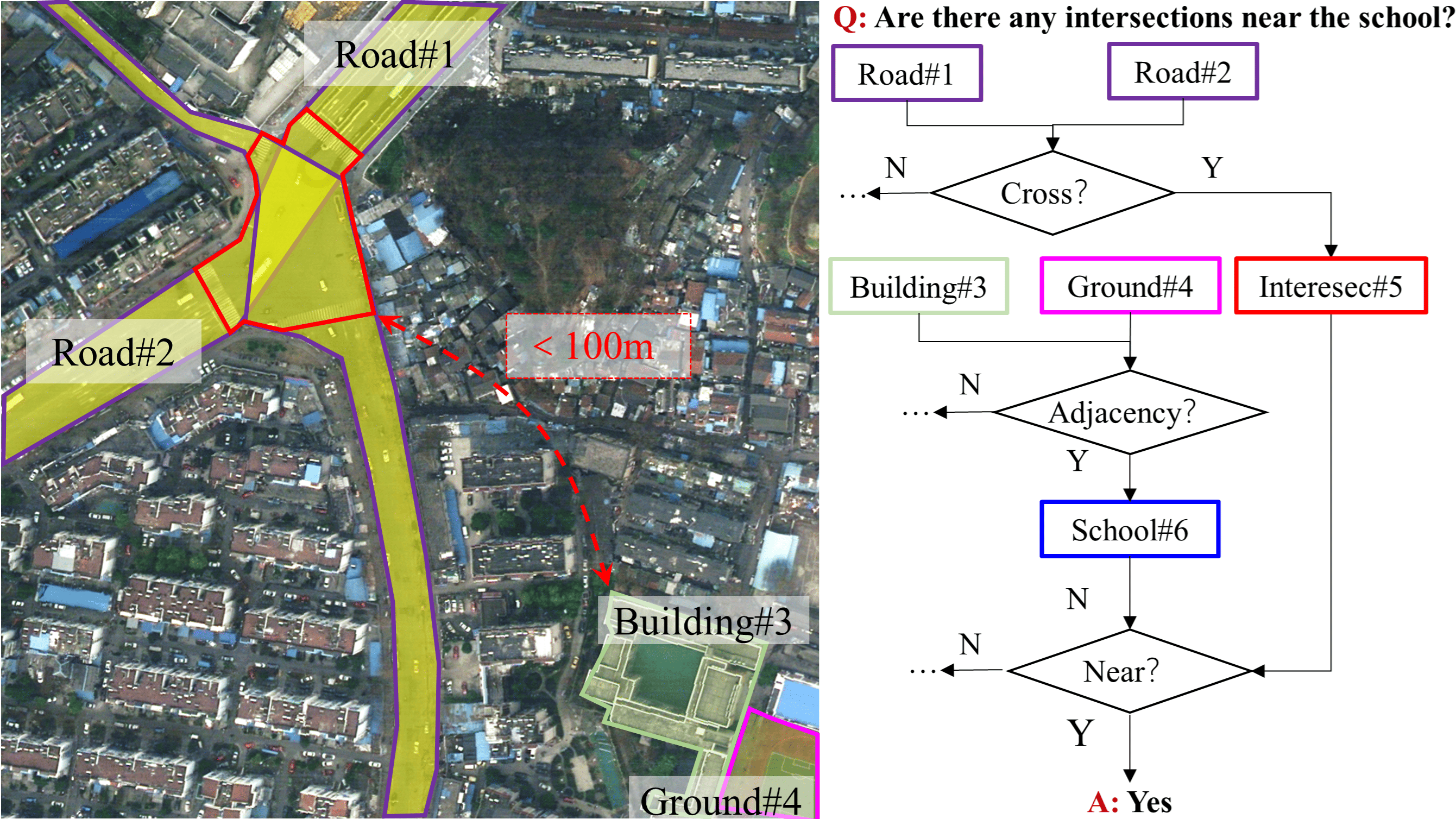}
  \vspace{-0.3cm}
  \caption{Answer annotation of the complex judgment question ``Are there any intersections near the school?''.
  }
  \label{fig:Annotation_RJ}
\vspace{-0.2cm}
\end{figure}

\begin{figure}[!hbt]
  \centering
  \includegraphics[width=1.0\linewidth]{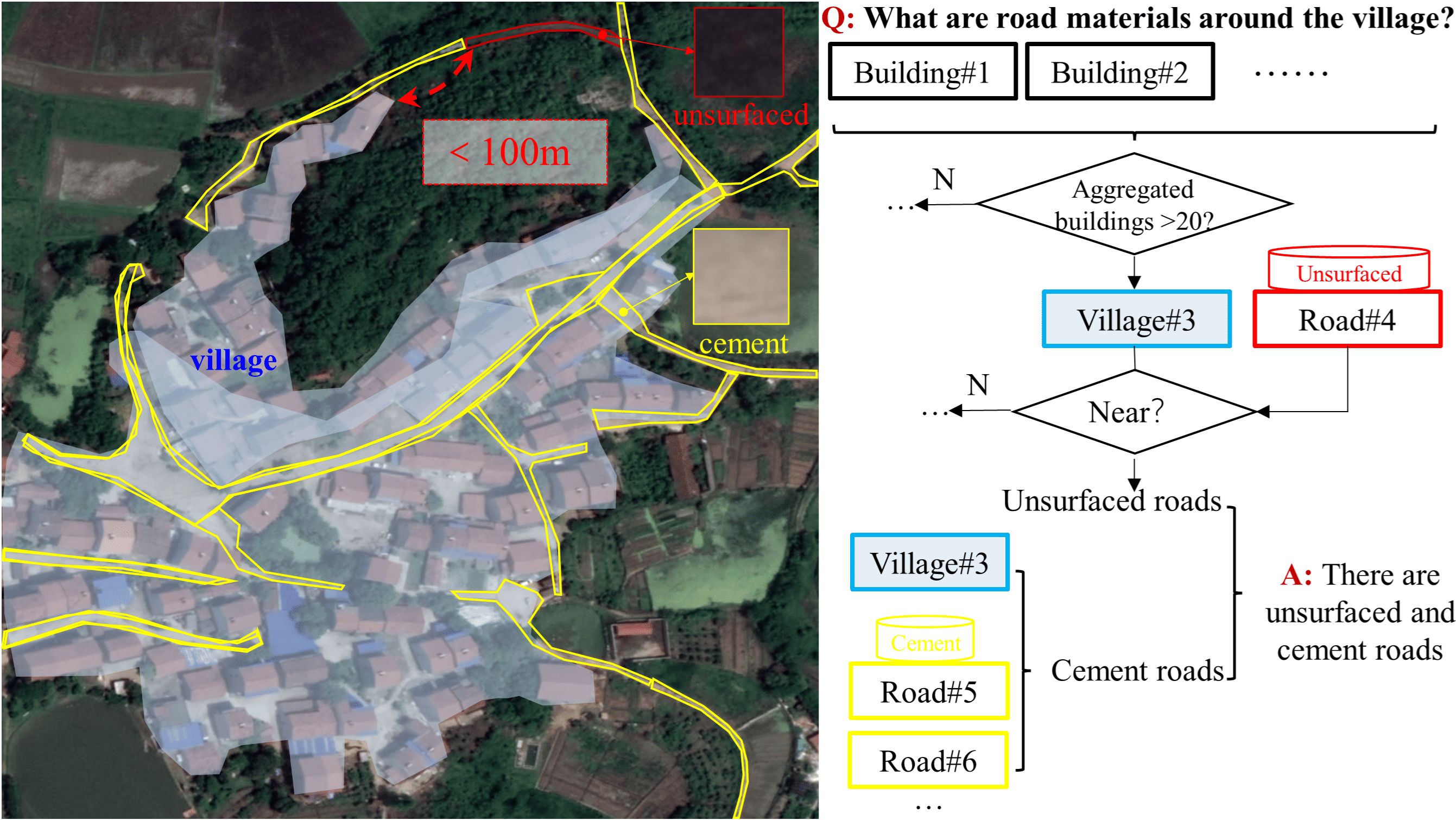}
  \vspace{-0.3cm}
  \caption{Answer annotation of the comprehensive analysis question ``What are the road materials around the village?''.
  }
  \label{fig:Annotation_CA}
  \vspace{-0.3cm}
\end{figure}

For the basic questions, the corresponding answers are automatically generated from the semantic masks. Given that each HSR image maintains a consistent spatial resolution of 0.3~m, 
the area estimation for basic objects is stratified into 10 discrete intervals, specifically $(x\%, x+10\%], x \in \{0, 10, ..., 90\}$.
To avoid ambiguous answers, we set a series of annotation guidelines for the complex questions.
The relational reasoning mainly includes the topologies, distances, sub-properties, conditional statistics, and directions.
Each step has fixed
thresholds and conditions. Using the
ArcGIS spatial analysis toolbox, professional annotators
can obtain a specific answer.

As for the complex judgment, the annotation procedure for ``Are there any intersections near the school?'' is depicted in
Fig.~\ref{fig:Annotation_RJ}.
By judging the topology,
the segmented Road\#1 and Road\#2 are crossed to first form Intersec\#5.
Furthermore, the teaching Building\#3 and Playground\#4 are adjacent and form the School\#6 scene.
Finally, the annotators utilized the ArcGIS toolbox to calculate the polygon-to-polygon distance between Intersec\#5 and School\#6, obtaining 94.8m.
Considering that the threshold of ``near'' is 100m, the complex judgment answer is ``Yes''.

As for the comprehensive analysis, Fig.~\ref{fig:Annotation_CA} shows
the annotation procedure for ``Are there any intersections near the school?''.
The annotators first searched for the village, which is formed of compact buildings (more than 20 buildings).
The aggregated buildings form a polygon of Village\#3, denoted by a light blue mask.
Most of the roads in this image are cement, but a small section of road has not yet been paved.
By judging the polygon-to-polygon distances,
all these roads are near to Village\#3.
Thus, the final answer can be obtained, i.e., ``There are unsurfaced and cement roads.''
Moreover, 
certain land-use categories such as commercial, industrial, and educational are identified with the aid of OSM data 
as supplementary information. 
\textit{EarthVLSet} deliberately excludes questions that could lead to ambiguity.
The criteria for the annotations in the dataset are defined as follows:
1) A distance of 100m is used to determine the proximity criterion labeled as ``near''.
2) An aggregation of more than 20 compact buildings is classified as a residential area.
3) Residential buildings display varied appearances and heights.
4) Commercial buildings are characterized by uniform appearances and orderly layouts.
5) Bodies of water exhibiting green algae and other types of floating vegetation are classified as eutrophic.
6) Some land-use types that reflect socioeconomic attributes, such as commercial and industrial areas, are identified using properties from OSM data.
7) A leaf area index (the ratio of vegetation area to total area) below 30\% in residential zones indicates a need for supplemental planting.
These guidelines ensure precise and consistent annotation within the dataset, enhancing the reliability of the research findings.

\begin{figure}[!hbt]
  \vspace{-0.1cm}
  \centering
  \includegraphics[width=1.0\linewidth]{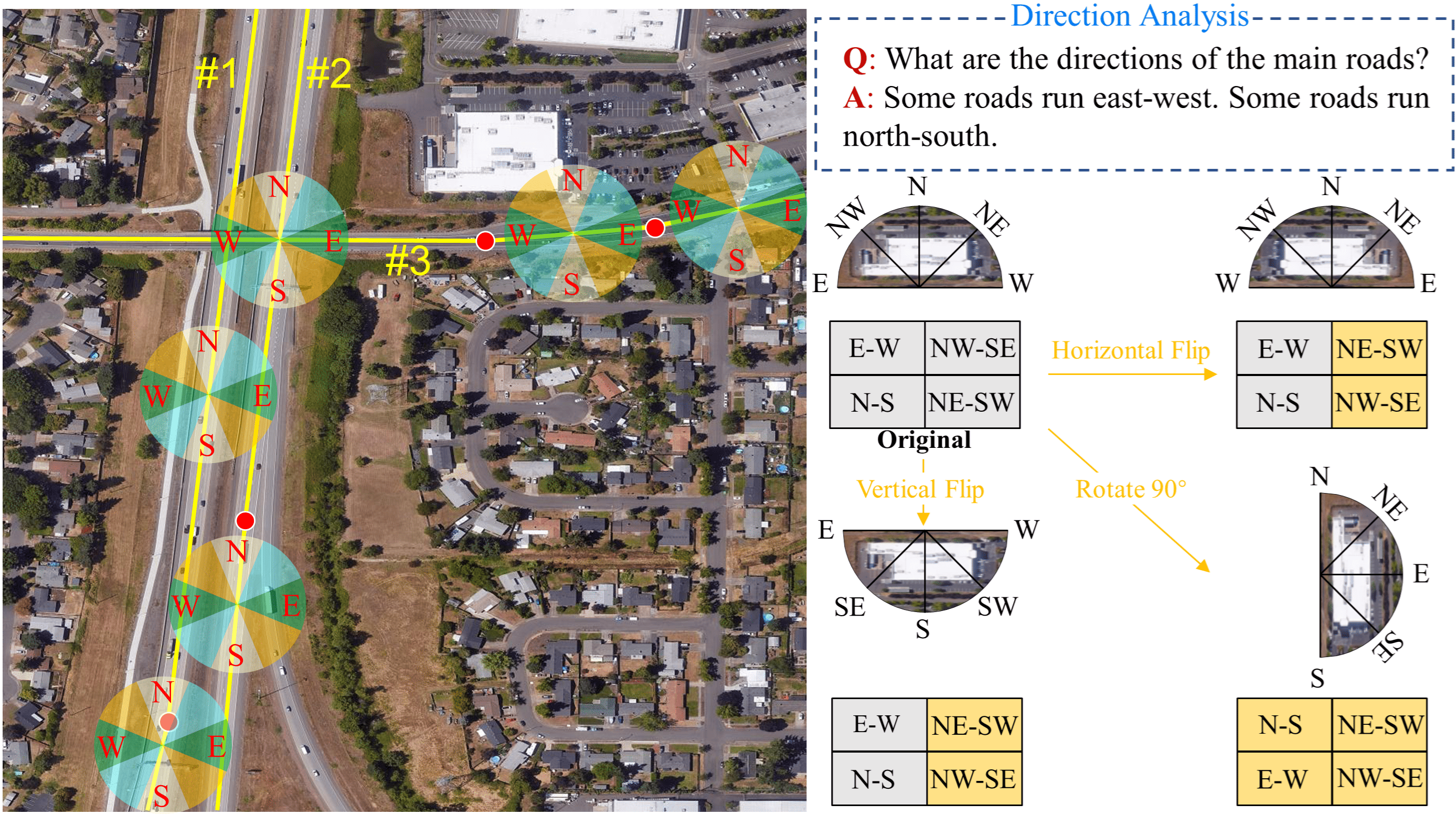}
  \vspace{-0.7cm}
  \caption{Answer annotation for a road direction analysis question. The geometric transformation in the data augmentation
  also affects the answer annotation.
  }\vspace{-0.4cm}
  \label{fig:Direction_Analysis}
\end{figure}
As for the newly added road direction analysis questions,
the annotation guideline is shown in Fig.~\ref{fig:Direction_Analysis}.
The direction candidates include ``east--west (E--W)'', ``north--south (N--S)'', ``northwest--southeast (NW--SE)'', and ``northeast--southwest (NE--SW)'', where
each direction covers the angle of $45^{\circ}$.
As the roads are annotated by polygons stored in shapefiles, 
we extracted their centerlines and obtained road segments according to the nodes.
For example, Road\#3 is divided by two red nodes, generating three straight segments.
The annotators needed to recognize the directions of each segment, and then gather them together.
Specifically, these three roads generate seven segments, and all the segments fall into the ``E--W'' or `'`N--S'' intervals.
According to the OSM road categories,
only main roads are considered for the direction analysis, while others (residential roads, tracks, cycleways, etc.) are filtered out.
Because the direction-sensitive words in the answer text are strongly related to the image geometric transformation, 
we set a series of guidelines to transform the answer text during the data augmentation. 
Specifically,
the horizontal and vertical flips affect the ``NW--SE'' and ``NE--SW'' while ``E--W'' and ``N--S'' remain unchanged.
In contrast, the $90^{\circ}$ rotation affects all the direction-sensitive words.
The code provides these answer augmentations for flip and rotation.
The implementation helps the direction analysis tasks align with the common training settings.

\subsection{Open-Ended VQA Data}
\label{sec:3:open}
\begin{figure}[!hbt]
  \vspace{-0.1cm}
  \centering
  \includegraphics[width=1.0\linewidth]{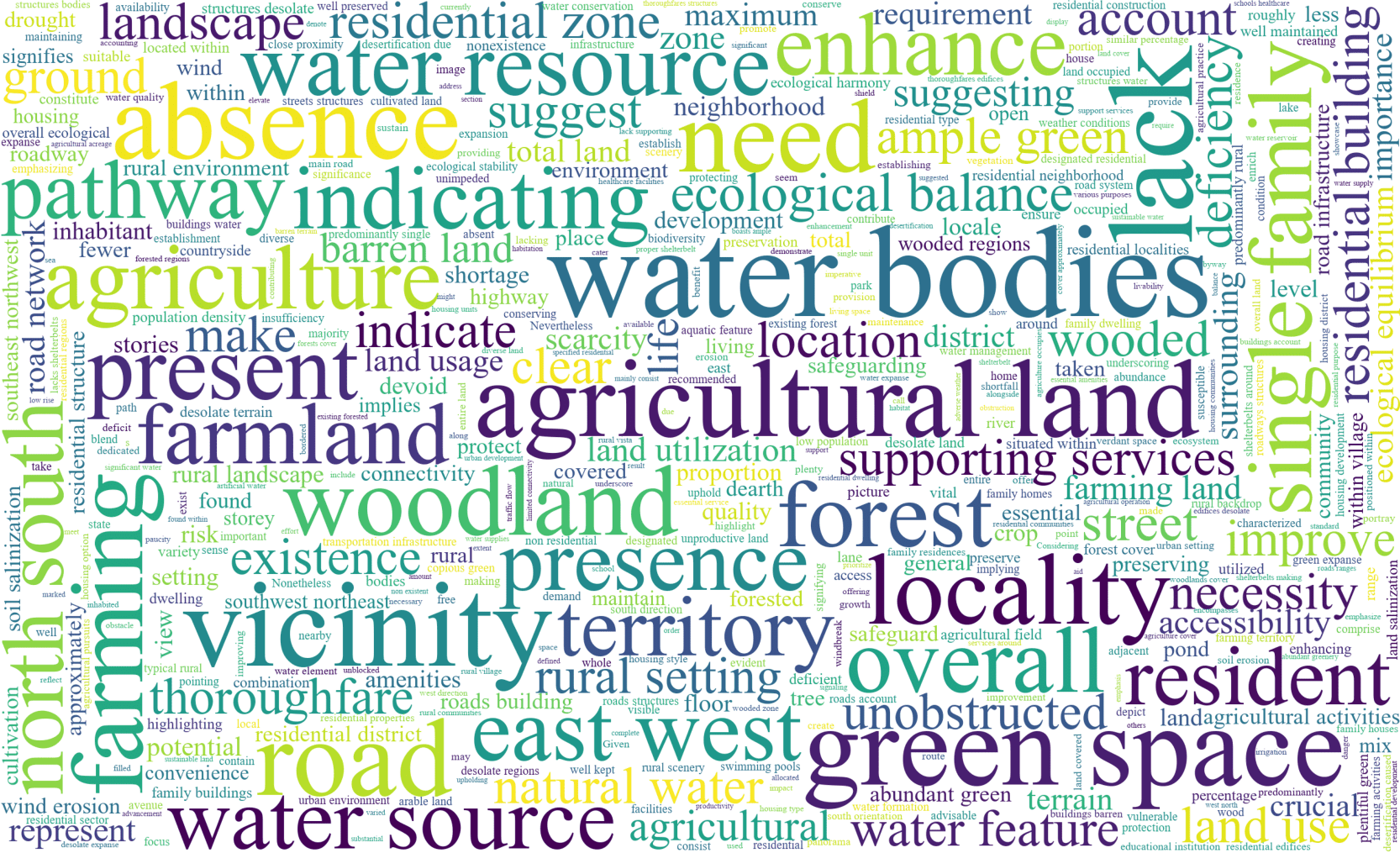}
  \vspace{-0.5cm}
  \caption{Word cloud visualization of the open-ended answers in the EarthVLSet. The word size is positively related to its frequency.
  }
  \label{fig:Word_Cloud}
  \vspace{-0.4cm}
\end{figure}

To achieve flexible answers, 238,454 open-ended QA pairs including sentences of variable lengths were constructed. 
The open-ended VQA tasks were designed for more comprehensive issues, i.e., scene descriptions and planning suggestions.
Based on urban planning concerns, seven questions for different topics were designed.
The question templates are ``Describe and give me some advice $\{topic\}$'', where $topic \in$ \{``about the development of residential buildings.'', ``about the development of living environments.'', ``about the greening renovation.'', ``based on the water situation.'',
``based on the traffic situation.'', ``about agriculture in this scene.'', ``about land cover objects in this scene.''\}.
Each question includes five answers with similar meanings.
Compared to multiple-choice data, the open-ended answers feature a richer vocabulary. The word cloud 
for the open-ended answers
is visualized in Fig.~\ref{fig:Word_Cloud}. 
The adjectives and nouns related to key geographical objects account for large proportions of the answers, posing challenges
for the complex modeling capabilities of language models. 


\noindent \textbf{Open-Ended QA Statistics.}
The number of samples for each open-ended question ranges from 27k to 40k, and the sufficient samples guarantee stable training for each type.
The question ``Describe and give me some advice about agriculture in this scene.'' has the fewest samples because only rural 
areas involve this question. As for the answer length, most of the open-ended answers are long because of the complex HSR remote-sensing scenes with lots of objects. Specifically, for ``Describe and give me some advice about land cover objects in this scene.'', the average length of the answers is about 52.76 words. 
Achieving these open-ended tasks requires not only accurate 
recognition of geographical objects but also effective causal language modeling for long sentences.

\begin{figure}[!hbt]
  \centering
  \includegraphics[width=1.0\linewidth]{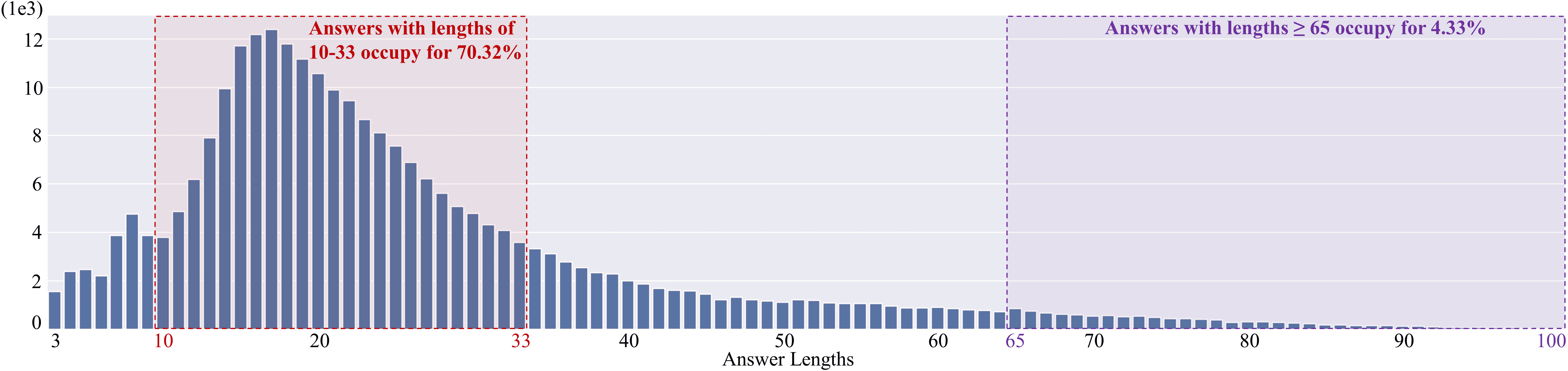}
  \vspace{-0.5cm}
  \caption{Answer distributions with different lengths in the open-ended data.
  The answers longer than 100 words are not included for simplification.
  Please zoom in for a better view.
  }\vspace{-0.2cm}
  \label{fig:len_dist}
\end{figure}

Fig.~\ref{fig:len_dist} shows the distributions of the answer lengths for the open-ended data.
It is clear that most answers (70.32\%) have word counts between 10 and 33, and the number of answers gradually decreases as the word count increases.
Specifically, the answers longer than 65 words only account for 4.33\%.
This long-tail distribution brings more challenges for long sentences, due to the fewer samples.

\begin{figure}[!hbt]
  \vspace{-0.3cm}
  \centering
  \includegraphics[width=1.0\linewidth]{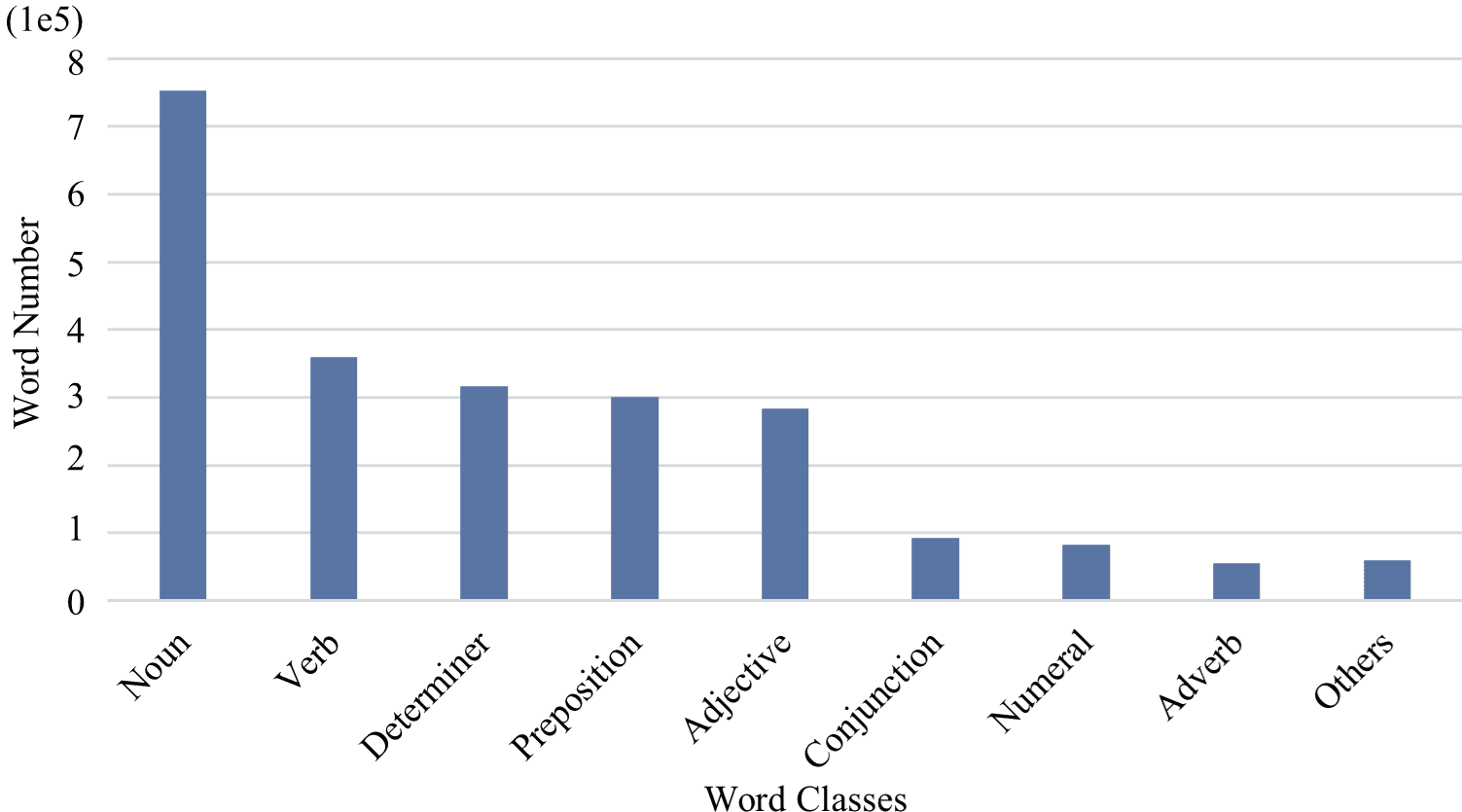}
  \vspace{-0.5cm}
  \caption{Word distributions with different classes. The nouns, verbs, determiners, prepositions, and adjectives have large proportions because the geographical objects are diverse in the complex HSR scenes.
  }
  \label{fig:pos_tag_dis}
  \vspace{-0.2cm}
\end{figure}

The word distributions for the different classes are shown in Fig.~\ref{fig:pos_tag_dis}.
It can be concluded that the answers in the proposed open-ended data are rich.
Sufficient nouns, verbs, determiners, prepositions, and adjectives
are required to describe the diverse geographical objects in HSR images clearly.
In contrast, adverbs occupy a smaller proportion because remote sensing images 
only describe objective geographical objects and do not include subjective emotions.
Compared to natural computer vision image captions, the 
words expressing degree, manner, etc. are relatively rare in remote sensing image captions.

\noindent \textbf{Annotation Guidelines and Quality Control.}
According to the open-ended questions, the annotators  
reorganized the information in the semantic masks and multi-choice answers to
generate the indefinite answers.
As for ``Describe and give me some advice about land cover objects in this scene.'',
the annotators first gathered the basic counting answers for each land-cover type for summarization.
After describing, advise is provided about the living environments with regard to natural, economic, and social situations. 
According to the multiple-choice annotations,
several rounds of inspection process were also conducted.  
Considering the linguistic diversity,
the manually annotated answers were augmented using synonymous sentence conversion
via GPT-4.
To this end, each open-ended question includes five similar and correct answers.

\section{Semantic-Guided EarthVLNet}
\begin{figure*}[hbt]
\centering
\includegraphics[width=1\linewidth]{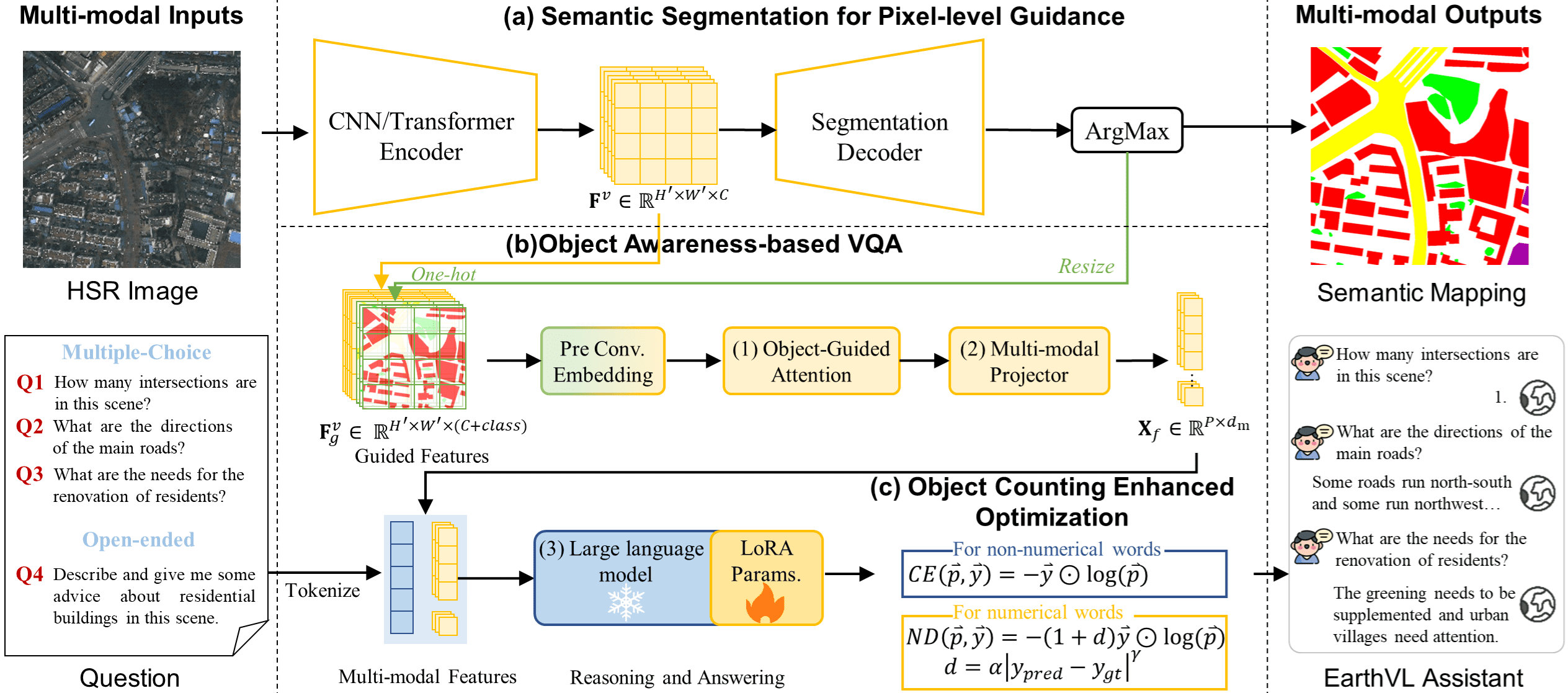}
\vspace{-0.3cm}
\caption{\textcolor{revision}{The proposed EarthVLNet includes a progressive learning architecture:
(a) Semantic Segmentation for Pixel-level Guidance; and (b) Object Awareness-based VQA. (c) Object Counting Enhanced Optimization improves the training of the word generation and object counting.}
} 
\vspace{-0.3cm}
\label{fig:framework}
\end{figure*}

As shown in Fig.~\ref{fig:framework},
\textit{EarthVLNet} includes two-stage training: 1) semantic segmentation network training for generating visual features and pseudo masks; and
2) semantic-guided VQA training for multi-modal reasoning and answering.

\subsection{Semantic Segmentation for Pixel-Level Guidance}
To handle HSR scenes with multiple objects, we innovatively employ a segmentation network for refined guidance. Given an input image
$\mathbf{I} \in \mathbb{R}^{H \times W \times 3}$, we extract visual features from the encoder outputs $\mathbf{F}^v \in \mathbb{R}^{H' \times W' \times C}$,
where $C$ represents the feature dimension and $H' = \frac{H}{32}, W'=\frac{W}{32}$ according to standard configurations.
We also use a pseudo-semantic output $\mathbf{M}^v \in \mathbb{R}^{H \times W}$ to enhance the object awareness.
In contrast to the traditional Faster-RCNN-based methods \cite{yu2019mcan, Anderson_2018_CVPR} that average box features into a single vector, the segmentation visual prompts retain the spatial locations and semantic details within objects. 
This improves the modeling of diverse compact geospatial objects.

\subsection{Object Awareness-Based LLM for VQA}
Guided by the questions and object semantics, the
object awareness based LLM reasons
visual cues for the final answers.
As shown in Fig.~\ref{fig:framework},
there are three components: 
1) object-guided attention (OGA) for object aggregation; 
2) multi-modal projector (MMP) for vision-language feature alignment; 
and 3) large language model (LLM) for relational reasoning and answer generation.
\begin{figure}[hbt]
  \centering
  \includegraphics[width=1\linewidth]{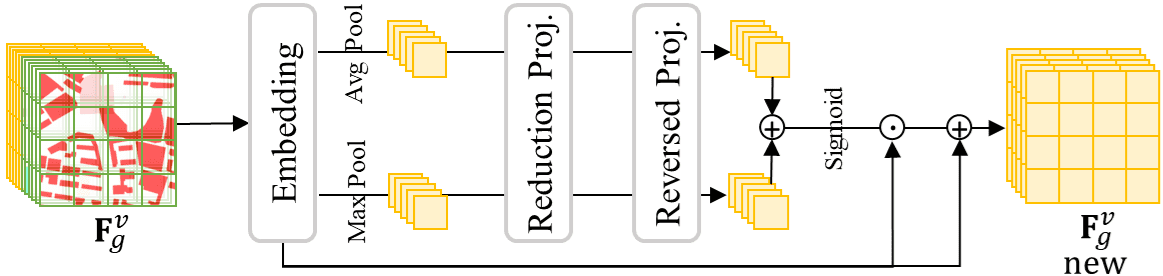}
  \vspace{-0.4cm}
  \caption{The object-guided attention includes max pooling and mean pooling for the channel-wise refinement, and the key object semantics are enhanced.} 
  \label{fig:OGA}
\end{figure}

\noindent \textbf{OGA for object aggregation.}
Because the segmentation output has explicit object details $\mathbf{M}^v$ (including categories and boundaries),
it is adopted to explicitly enhance the visual features.
As shown in Fig.~\ref{fig:OGA},
OGA is proposed to dynamically weight $\mathbf{F}^v$ and $\mathbf{M}^v$ from the channel dimension.
Using nearest-neighbor interpolation,
$\mathbf{M}^v$ is first resized into the same size as $\mathbf{F}^v$.
One-hot encoding followed by a pre-convolutional embedding effectively serializes the object semantics. The embedding contains a 3$\times$3 convolution, batch normalization, and a ReLU. 
They are concatenated to obtain object-guided features $\mathbf{F}^v_g$ as inputs for OGA.
Inspired by previous work \cite{woo2018cbam}, OGA consists of spatial and dimensional refinement.
The reduction and reverse projections further refine the features dimensionally.
After activation, we use 
the refined features to calibrate the subspaces of $\mathbf{F}^v_g$ from the channel dimension. 

\noindent \textbf{MMP for Feature Alignment.}
To align the visual features with the language features,
an MMP is adopted \cite{liu2024llavanext}, including two linear layers with a GELU activation inserted.
Through this non-linear projection, the pixel-level guided visual features can be effectively
refined before the multi-modal feature fusion.

\noindent \textbf{LLM for Relational Reasoning.}
To model complex relations and generate
diverse indefinite-length answers, we transform the traditional classification decoder with an LLM.
As for question inputs, the language tokenizer transforms the text into language features.
After the concatenation of the vision and language features, the multi-modal features are then processed with 
an LLM. The LLM aims to reason key object relations and generate the final answers.
Due to the large model size of the LLM, fine-tuning each weight in the colossal model is impractical.
As a parameter-efficient fine-tuning approach,
the LoRA freezes the pre-trained weights and fine-tunes the few injected adapters\cite{hu2021lora}. 
The LoRA ensures faster convergence and maintains the original knowledge learned from the
generic natural language instructions. 
During the adaptation, the LLM gradually fits the HSR scenes with the land-cover semantics and relations.
\begin{figure*}[hbt]
  \centering
  \includegraphics[width=0.9\linewidth]{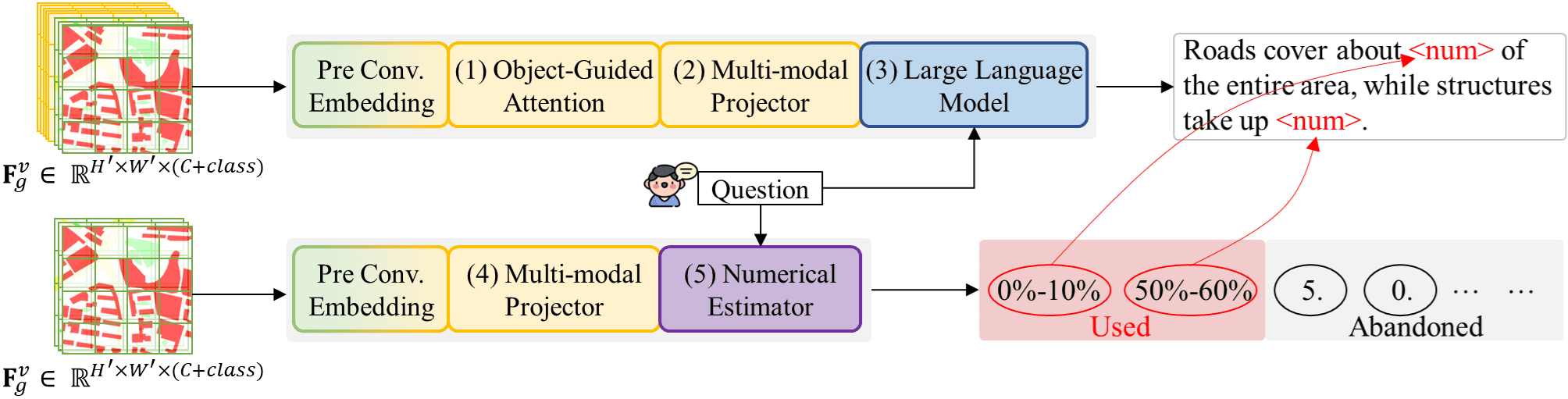}
  \caption{The object counting enhanced optimization separately models the conditional generation and counting estimation.
  The conditional generation (\textbf{Top}) only considers the non-numerical words in the answers, and
  ``\textcolor{red}{$\textless$num$\textgreater$}'' refers to the numerical placeholders.
  The object counting estimation (\textbf{Bottom}) aims to obtain accurate numbers to fill out the answers.
  }
  \label{fig:numerical_estimator}
  \vspace{-0.3cm}
\end{figure*}

\subsection{Object Counting Enhanced Optimization}
VQA tasks include both classification and regression (object counting) questions. However, the existing methods regard them as a multi-classification task, which is processed with cross-entropy (CE) loss.
Eq~\eqref{eq:ce} indicates that CE loss is insensitive to the distance between the predicted value and the true value, and is therefore not suitable for the regression task.
\begin{equation}
CE(\vec{p} ,\vec{y})  = -\vec{y} \odot log(\vec{p}) =\sum^{class}_{i=1} -y_i log(p_i) \label{eq:ce}
\end{equation}
where $\vec{y}$ specifies the one-hot encoded ground truth,
$\vec{p}$ denotes the predicted probabilities, and
$i$ represents the class index for each answer.
To introduce a difference penalty for the regression task, we add a modulating factor $d = \alpha |\mathbf{y}_{diff}|^{\gamma} =\alpha |\mathbf{y}_{pr} - \mathbf{y}_{gt}|^{\gamma}$
to the CE loss. $\mathbf{y}_{pr}$ and $\mathbf{y}_{gt}$ represent the predicted and ground truth number, respectively. $\alpha \geq 0$ and $\gamma \geq 0$ are tunable distance awareness factors.
$d$ represents the distance penalty $d \propto \mathbf{y}_{diff}$. 
Finally, the numerical difference (ND) loss is designed as follows:
\begin{equation}
\begin{split}
ND(\vec{p} ,\vec{y})  &= -(1 + d) \vec{y} \odot log(\vec{p}) \\
&= -(1 + \alpha |\mathbf{y}_{diff}|^{\gamma}) \vec{y} \odot log(\vec{p}) \\
&=  -(1 + \alpha |\mathbf{y}_{pr} - \mathbf{y}_{gt}|^{\gamma}) \sum^{class}_{i=1} y_i log(p_i)
\end{split}\label{eq:nl}
\end{equation}

The ND loss unifies the classification and regression objectives into one optimization framework.
$\alpha$ controls the overall penalty for the regression tasks, compared to the classification tasks.
$\gamma$ determines the sensitivity of the regression penalty to numerical differences.
We plotted the relationship between penalty $d$ and distance difference $\mathbf{y}_{diff}$.
As $\alpha$ increases, the overall penalty increases, meaning that the optimization focuses more
on regression tasks.
With $\alpha=0$, the ND loss degenerates into the original CE loss and the penalty is constant ($d = 0$ when $|\mathbf{y}_{diff}| \in [0, +\infty)$).
The sensitivity of the regression penalty increases as $\gamma$ increases,
and when $\gamma > 1$, the penalty curve changes from concave to convex.

Compared to generic VQA images,
remote sensing scenes contain various geospatial objects, so that the counting estimation is more challenging.
We model the conditional generation and object counting processes separately (Fig.~\ref{fig:numerical_estimator}) because 
different tasks will be mutually exclusive during the optimization \cite{xiao2018unified}.
As for supervised labels, the answers including numerical words are masked with placeholders ($\textless$num$\textgreater$), and the 
original numbers are extracted to form a sequence.
As for modeling, the LLM receives the fusion of the semantic features and pseudo masks as input, generating the non-numerical words. 
Because the pseudo masks
explicitly include the locations and categories of the geospatial objects,
the numerical estimator utilizes these for effective object counting. 
In essence, the LLM generates the non-numerical words and predicts the positions of the numerical words. 
The numerical estimator focuses on statistical analysis and object counting.
In implementation, the numerical estimator is constructed based on the stacked Transformer blocks.
Each Transformer block includes a self-attention and a feed-forward network.



\section{Experiments}
As \textit{EarthVLSet} promotes both land-cover semantic segmentation and 
VQA, we performed comprehensive benchmarkings on three tasks, exploring 
the relations between vision and language data in Earth observation scenes.
\subsection{EarthVLSet Division}
As for the dataset division, following the EarthVQA dataset \cite{wang2024earthvqa},
we split the HSR images based on geographical isolation laws. 
The \texttt{Train} set includes 17 areas covering Markov, Louisville, and Eugene in America;
Singapore; Casablanca in Morocco; Paris in France; Arabia in Riyadh; Port Hedland in Australia; Damascus in Syria; and
Nanjing (Qixia, Gulou, Qinhuai, Pukou Gaochun, Lishui) and
Wuhan (Jianghan, Jiangxia) in China.
The \texttt{Val} set includes nine areas covering Callao in Peru; New South Wales in Australia; Rotterdam in the Netherlands; Engels in Russia; Sao Paulo in Brazil;
and Nanjing (Yuhuatai, Liuhe), Changzhou (Jintan), and Wuhan (Huangpi) in China.
The \texttt{Test} set includes 12 areas covering
Pake and New York in America; Hakodate in Japan; Cairo in Egypt; Bangkok in Thailand; Rome in Italy; and
Nanjing (Jianye, Jiangning), Changzhou (Wujin, Liyang, Xinbei), and Wuhan (Wuchang) in China.
Because the cities of Nanjing, Changzhou, and Wuhan include more than one sampled region, we specify the districts in parentheses.
The \texttt{train} set contains 5,260 images, 227,030 multiple-choice QA pairs, and 135,352 open-ended QA pairs.
The \texttt{val} set contains 2,699 images, 116,973 multiple-choice QA pairs, and 38,764 open-ended QA pairs.
The \texttt{test} set contains 3,336 images, 152,019 multiple-choice QA pairs, and 91,461 open-ended QA pairs.
Each set has sufficient urban and rural samples, ensuring diversity of the training and evaluation.
\begin{table*}[!hbt]
  \caption{Land-cover semantic segmentation benchmarks on ConvNet-based and Transformer-based methods} \label{tab:comp_res}
  \resizebox{1.0\linewidth}{!}{
  \begin{tabular}{l|l|c|cccccccc|cc}
  \hline
  \multirow{2}{*}{Method} & \multirow{2}{*}{Backbone} & \multirow{2}{*}{$\uparrow$mIoU(\%)} & \multicolumn{8}{c|}{$\uparrow$IoU per category (\%)}              & \multirow{2}{*}{Params}  &  \multirow{2}{*}{FLOPs}   \\ \cline{4-11}
          &         &         & Background & Building & Road  & Water & Barren & Forest & Agriculture & Playground & &  \\ \hline
  \multicolumn{2}{l|}{\rsb \textit{ConvNet-based}}           &         &    &    &   &   &    &    &     &    &     &   \\
  FCN8S\cite{long2015fully}   & VGG16       & 47.43       & 36.12  & 51.12  & 41.57 & 74.64 & 26.89  & 56.86  & 57.64   & 34.59  & 15.3M  & 180.7G \\
  UNet\cite{diakogiannis2020resunet}       &ResNet50       & 51.74	&39.35	&56.28	&45.74	&79.35	&26.44	&58.42	&61.49	&46.82  & 32.5M & 96.6G \\
  UNet++\cite{zhou2018unet++}       & ResNet50  & 52.54       & 39.09  & 57.30  & 49.72 & 80.20 & 26.41  & 58.01  & 62.51   & 47.10  & 48.9M  & 518.3G \\
  DeepLabV3+\cite{chen2018encoder}     & ResNet50      & 50.88 &38.84 &54.64 &47.11 &78.82 &25.64 &56.94 &60.00 &45.06  & 26.6M  & 82.7G \\
  PAN\cite{li2018pyramid}       & ResNet50      & 51.26 &40.17& 55.50 &46.63 &78.77& 25.44 &58.43 &62.34 &42.78  & 24.2M  & 78.3G \\
  PSPNet\cite{PSPNet}       & ResNet50      & 51.53& 39.70& 55.95& 46.84& 80.20 &22.09 &58.27 &62.73 &46.46  & 53.3M  & 453.3G \\
  LinkNet\cite{chaurasia2017linknet}      & ResNet50  & 51.02  &38.21	&54.88	&47.45	&79.09	&24.95	&59.09	&60.12	&44.38  & 31.1M  & 65.9G  \\
  FarSeg\cite{farsegpp}       & ResNet50      & 52.66 & 39.87 & 57.58&  49.52&  80.27&  24.76 & 58.94& 62.38 & 47.99  & 31.3M  & 105.8G \\
  FactSeg\cite{FactSeg}      & ResNet50      & 51.96  & 38.95 &  55.86 &  49.14 & 79.65 &  24.35 &  58.77 &  61.18 &  47.78 & 33.4M  & 99.4G \\
  HRNet\cite{wang2020deep}      & W32    & 53.40& 39.85 &58.92 &51.65& 81.19 &27.69 &60.37 &62.14 &45.40  & 29.5M  & 102.0G  \\
  Bi-FPN\cite{tan2020efficientdet}  & ResNet50  & 52.28 &39.46 &57.50 &49.02 &79.37 &24.45 &58.36 &61.62 &48.46  & 28.3M  & 81.7G  \\
  Semantic-FPN\cite{SFPN}     & ResNet50  & 52.01 &39.19 &56.00& 49.05 &79.90& 25.99 &57.82& 61.80& 46.29  & 28.4M   & 44.2G \\
  Semantic-FPN    & ConvX-T\cite{liu2022convnet}  & 53.56 & 40.97 & 59.54  &49.85 & 81.03& 24.14& \textbf{60.83}& 65.67& 46.48  & 32.1M   & 44.7G  \\ 
  UperNet\cite{xiao2018unified}     & ConvX-T  & 53.45 &40.73 &58.65& 50.40& 80.67 &26.97 &59.51& 64.82& 45.89 &59.2M   & 525.7G  \\
  SegNext\cite{guo2022segnext}   & MSCAN-B   & \underline{54.94}& \textbf{42.32}& 60.16 &\textbf{53.01} &81.17 &27.25 &59.71 &\underline{66.73} &49.13 &26.7M   & 64.7G  \\ \hline
  \multicolumn{2}{l|}{\gb \textit{Transformer-based}}           &    &    &   &   &    &    &     &    &     &   \\
  MobileVIT\cite{MobileViT}    & Mob-S       & 47.75&37.98&53.35&46.63&79.58&31.13&57.94&61.42&13.96  & 6.3M   & 40.1G     \\   
  SegFormer\cite{xie2021segformer}    & MiT-B2      & 54.34 &41.03 &\textbf{60.83}& 50.89& \textbf{81.76}& 29.24& 59.74 &64.93& 46.30 &24.7M  & 67.7G  \\ 
  Mask2Former\cite{cheng2022masked}    & Swin-T     & 53.69 &40.23 &57.73 &50.45& 79.59& 28.72 &58.34& 63.71& \textbf{50.78} &47.3M   & 139.6G  \\ 
  Semantic-FPN     & Swin-T\cite{liu2021swin}       & 54.42 &40.32 &58.50 &49.34& \underline{81.35} &\underline{31.35} &60.17& 64.61& \underline{49.71} &31.8M     & 46.9G   \\ 
  \textcolor{revision}{TransUNet\cite{chen2021transunet}} & \textcolor{revision}{R50-ViT-B/16} & \textcolor{revision}{\textbf{55.00}} & \textcolor{revision}{\underline{41.44}} & \textcolor{revision}{\underline{60.61}} & \textcolor{revision}{\underline{51.77}} & \textcolor{revision}{80.32} & \textcolor{revision}{29.36} & \textcolor{revision}{\underline{60.44}} & \textcolor{revision}{\textbf{67.34}} & \textcolor{revision}{48.74} & \textcolor{revision}{105.91M} & \textcolor{revision}{158.6G} \\
  UperNet    & Swin-T       & 53.96 &40.68 &58.63 &48.78& 79.94& \textbf{31.58}& 58.94 &64.28 &48.85  & 32.1M   & 528.0G  \\ \hline
  \end{tabular}}
\end{table*}

\subsection{Land-cover Semantic Segmentation}
\textbf{Implementation Details.}
The semantic segmentation networks were implemented under the PyTorch framework, and the
experiments were conducted using two 24GB RTX 4090 GPUs.
We used the AdamW optimizer with $\beta=(0.9, 0.999)$ and a weight decay of 0.05. 
The base learning rate was set to 1e-4 and controlled by a ``poly'' schedule with a power of 0.9.
The batch size was 16, and all the models were trained for 30k steps.
As for data augmentation, the images were first randomly scaled with ratios of \{0.5, 0.75, 1.0, 1.25, 1.5, 1.75, 2.0\} and then randomly cropped into 512 $\times$ 512 patches.
Random flipping, rotation, and color jitter were also applied for the data augmentation.

\noindent \textbf{Comparative Results.}
To recognize the object locations and categories accurately,
we evaluated 18 advanced semantic segmentation methods, involving both general computer vision and remote sensing methods. 
The comparative results provided in 
Tab.~\ref{tab:comp_res} indicate that the different methods 
show large differences in accuracy.
Thanks to the diversity of \textit{EarthVLSet},
the generalizability of semantic segmentation methods can be effectively distinguished.
The lightweight and traditional architectures with shallow layers, i.e., MobileVIT and FCN8S,
fail to achieve satisfactory performances, due to the lack of representational ability.
As for HSR land-cover mapping tasks, the decoder is also important to restore the details of multi-scale objects.
Equipped with the ResNet50, 
UNet++ outperforms DeepLabV3+ in system-level accuracy by 1.66\%.
This is because the decoder of UNet effectively reuses the high resolution features in the encoder, which contributes to the restoration of small objects.
In conclusion, a well-established HSR semantic segmentation architecture is intended to grasp the abilities of 
multi-scale context interaction and refined detail recovery.

\begin{figure}[hbt]
  \centering
  \includegraphics[width=1\linewidth]{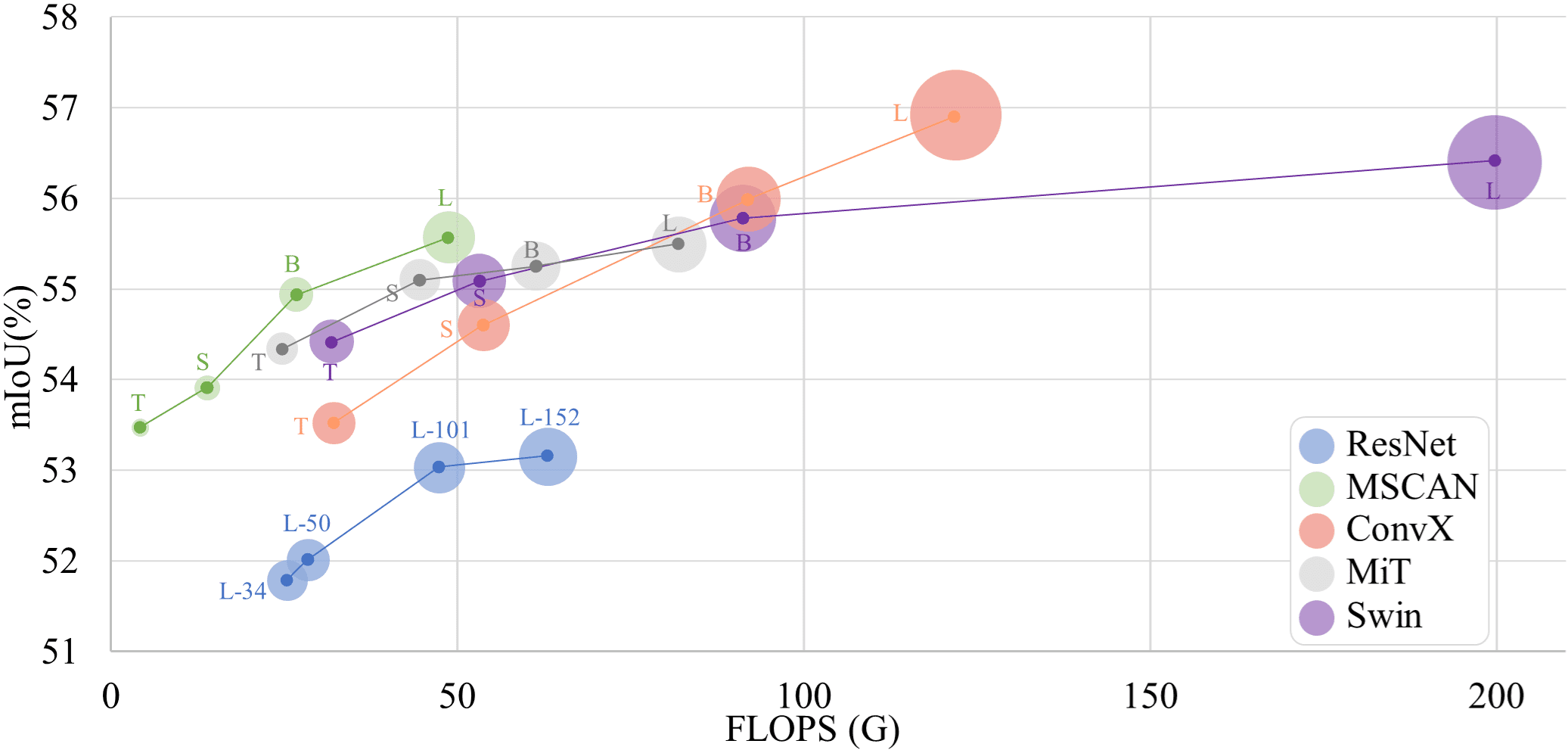}
  \vspace{-0.3cm}
  \caption{The semantic segmentation results of different vision backbones.
  L-34, L-50, L-101, and L-152 denote ResNet34, ResNet50, ResNet101, and ResNet152.
  T, S, B, and L denote Tiny, Small, Base and Large sizes. 
  } 
  \label{fig:backbone-exp}
  \vspace{-0.5cm}
\end{figure}

\noindent \textbf{Various Vision Encoders.}
As different encoders have great effects on the segmentation results,
we scaled up the backbones and kept the same pyramid feature decoder in Semantic-FPN.
Fig.~\ref{fig:backbone-exp} shows the comparative results using the 
ResNets, ConvNeXts, MSCANs, Swin-Transformers, and MiTs at different model scales.
At similar model sizes, the ResNets achieve lower performances compared to others, showing limited
generalizability for global-scale mapping.
Swin-Transformers and ConvNeXts model the multi-scale features
from different aspects, i.e., attention and convolution.
Both models are suitable for the \textit{EarthVLSet} semantic segmentation tasks.
MSCANs and MiTs are originally proposed as lightweight architectures, and can also
achieve competitive results. They can serve as effective solutions when faced with limited resources and time.
As for the basis of the downstream VQA tasks, more accurate semantic features contribute to better VQA performances \cite{wang2024earthvqa}. 
Semantic-FPN (with ConvNeXt-L) was chosen as the vision encoder by default for our VQA tasks.

\begin{table*}[!hbt]
  \centering
  \caption{\textcolor{revision}{Multiple-choice VQA benchmarks for the general-purpose and remote sensing tailored VLMs}}
  \label{tab:vqa_comp_res}
  \resizebox{1.0\linewidth}{!}{
  \begin{tabular}{l|c|c|c|cccccccc|c|ccc}
  \hline
  \multirow{2}{*}{Method}  & \multirow{2}{*}{Seg} & \multirow{2}{*}{Params} & \multirow{2}{*}{$\uparrow$OA(\%)} &
  \multicolumn{8}{c|}{$\uparrow$OA per class(\%)}  & \multirow{2}{*}{$\downarrow$RMSE} &
  \multicolumn{3}{c}{$\downarrow$RMSE per class} \\
  \cline{5-12} \cline{14-16}
               &      &  &     & BJ  & CJ  & BC  & CC  & AE & DisA  & DirA  & CA  &         & BC    & CC    & CA   \\ \hline

  \multicolumn{2}{l|}{\rsb \textit{Classification-based}} &     &   &   &   &   &  & & &   &         &   &   &   \\
  MAC\cite{hudson2018compositional}  & $\times$   & 49.9M & 73.89   & 78.53 & 84.49 & 73.36 & 59.29 & 54.40  & 51.47 & 32.94 & 57.92 & 3.379 & 1.818 & 6.008 & 9.499 \\
  RSVQA\cite{lobry2020rsvqa}     & $\times$   & 34.9M & 73.45   & 79.22 & 85.52 & 71.61 & 68.16 & 54.94 & 26.07 & 26.71 & 50.42 & 4.012 & 2.106 & 7.182 & 11.581 \\
  RSIVQA\cite{zheng2021mutual}   & $\times$   & 72.5M & 77.79   & 84.52 & 86.12 & 76.71 & 71.22 & 63.69 & 45.24 & 39.52 & 61.34 & 3.381 & 1.302 & 6.144 & 10.290 \\
  SAN\cite{yang2016stacked}    & $\times$   & 37.3M & 77.24   & 82.07 & 86.29 & 75.13 & 71.46 & 62.81 & 49.11 & 30.05 & 60.54 & 3.326 & 1.479 & 5.948 & 10.506 \\
  BAN\cite{kim2018bilinear}    & $\checkmark$ & 30.2M & 78.97   & 88.55 & 86.68 & 78.45 & 72.04 & 66.34 & 50.38 & 33.36 & 63.49 & 2.864 & 1.291 & 4.953 & 8.906 \\
  MCAN\cite{yu2019mcan}      & $\checkmark$ & 17.7M & 79.15   & 88.19 & 86.93 & 80.10  & 72.88 & \underline{67.14} & 51.38 & 39.61 & 63.87 & 2.577 & 0.984 & 4.907 & 8.438  \\
  BUTD\cite{Anderson_2018_CVPR}    & $\checkmark$ & 12.3M & 79.26   & 87.15 & 86.64 & 78.35 & 72.38 & 66.53 & 48.28 & 36.43 & 63.21 & 2.568 & 0.993 & 4.826 & 8.224 \\
  LXMERT\cite{tan2019lxmert}     & $\checkmark$ & 87.6M & 79.27   & 88.32 & 86.07 & 79.98 & 72.57 & 67.02 & 50.33 & 37.59 & 63.97 & 2.594 & 1.088 & 4.894 & 7.981  \\
  D-VQA\cite{wen2021debiased}    & $\checkmark$ & 17.6M & 77.80   & 85.39 & 86.23 & 78.37 & 71.84 & 59.58 & 47.84 & 33.05 & 60.91 & 2.848 & 1.122 & 5.414 & 8.273  \\ \hline

  \multicolumn{2}{l|}{\gb \textit{Generation-based}}  &     &   &   &   &   &  & & &   &         &   &   &   \\
  ALBEF\cite{li2021align}      & $\times$   & 290.6M & 74.21 & 81.42 & 85.19 & 76.29 & 62.92 & 48.94  & 44.31 & 29.93 & 51.95 & 2.686 & 1.158 & 4.942 & 8.094 \\
  BLIP-2\cite{BLIP2}       & $\times$   & 3.9B & 69.43 & 83.38 & 85.22 & 71.61 & 67.50 & 39.68 & 16.24 & 15.73 & 26.10 & 3.726 & 2.106 & 6.250 & 11.581 \\
  InstructBLIP\cite{liu2023improved}  & $\times$   & 4.0B & 78.04 & 87.57 & 86.21 & 76.72 & 72.00 & 64.31 & 43.75 & 39.17 & 55.26 & 2.758 & 0.980 & 5.309 & 9.138 \\
  LLaVANeXT\cite{liu2024llavanext}  & $\times$   & 7.2B & 79.32   & 87.92 & 86.71 & 78.88 & 72.74 & 66.64 & 51.87 & 38.33 & 64.33 & 2.721 & 1.133 & 4.901 & 8.644 \\
  \textcolor{revision}{LLaVA-OV}\cite{lillava}  & $\times$   & 8.0B &
  \underline{80.42} & \underline{89.06} & 87.53 & 80.12 & \underline{73.57} & 67.09 & 49.88 & \textbf{43.24} & 63.14 & 2.540 & 0.967 & 4.672 & 8.524 \\
  \textcolor{revision}{ViP-LLaVA}\cite{cai2024vip}  & $\times$   & 7.2B &
  79.78 & 88.57 & \underline{87.67} & 79.16 & 73.13 & 67.11 & \underline{52.36} & \underline{42.24} & 62.96 & 2.574 & 0.985 & 4.902 & 8.434 \\
  GeoChat\cite{kuckreja2023geochat} & $\times$  & 7.2B & 79.13   & 88.40 & 86.41 & 78.92 & 72.67 & 66.14 & 42.59 & 38.14 & 63.77 & 2.766 & 1.253 & 5.038 & 8.858 \\ \hline

  GPT-4o\cite{achiam2023gpt} & $\times$  & - & 61.15 & 84.55 & 70.63 & 48.83 & 36.74 & 50.44 & 28.94 & 15.05 & 28.95 & 3.507 & 2.331 & 5.707 & 12.56 \\
  Claude3 Opus\cite{gpto1} & $\times$  & - & 63.78 & 84.22 & 74.52 & 50.08 & 37.09 & 57.02 & 23.93 & 16.39 & 34.81 & 3.248 & 2.329 & \underline{4.506} & 10.76 \\ \hline

  \rsb \textcolor{revision}{SOBA~\cite{wang2024earthvqa}} & $\checkmark$ & 19.9M  &
  79.95 & 88.60 & 86.38 & \underline{80.23} & 73.18 & \textbf{67.21} & 51.10 & 39.70 & \underline{64.66} &
  \underline{2.482} & \textbf{0.905} & 4.654 & \underline{7.858} \\
  \gb EarthVLNet w.o. seg & $\times$ & 6.9B  &
  79.63 & 88.41 & 86.25 & 79.02 & 72.88 & 66.57 & 52.30 & 39.40 & 63.43 &
  2.636 & 1.013 & 4.808 & 8.042 \\
  \gb EarthVLNet (ours)   & $\checkmark$ & 6.9B  &
  \textbf{81.06} & \textbf{89.24} & \textbf{88.01} & \textbf{81.16} & \textbf{74.83} & 66.20 & \textbf{58.51} & 42.03 & \textbf{64.90} &
  \textbf{2.340} & \underline{0.908} & \textbf{4.341} & \textbf{7.141} \\
  \hline
  \end{tabular}}
\end{table*}

\subsection{Multiple-Choice Visual Question Answering} \label{sec:mc}
To evaluate the relational reasoning ability of the proposed \textit{EarthVLNet}, 
we first performed comparative experiments on the multiple-choice VQA task.  
We chose 13 advanced VQA methods covering both general multi-modal learning and remote sensing fields
for comparison.
GPT-4o and Claude were selected to show the zero-shot abilities  of general models.
Following the common settings\cite{yu2019mcan,wang2024earthvqa}, we adopted the classification accuracy and root-mean-square error (RMSE) as the evaluation metrics, with the
RMSE used to evaluate the counting tasks.

\noindent \textbf{Implementation Details.}
As for VQA methods that require semantic guidance,
the visual features of Semantic-FPN (ConvX-L) were adopted fairly.
All the VQA models were trained for
40k steps with a batch size of 16.
As for the large VLMs, BLIP-2 and InstructBLIP trained Q-Former following their original settings. The
vision encoder adopted ViT-g/14 and the language decoder leveraged FlanT5XL.
To scale up the language decoder,
LLaVaNeXt and GeoChat utilized Vicuna-7B for the LoRA fine-tuning.
The hyperparameters of LoRA were set as $r=64$ and $\alpha=16$.
As for \textit{EarthVLNet}, the LLM utilized Vicuna-7B and the counting part included 
three-layer Transformer blocks with a hidden size of 384.
We used the Adam solver with
$\beta_1$ = 0.9 and $\beta_2$ = 0.999. The initial learning rate was set
to 2e-4, and a ``poly'' schedule with a power of 0.9 was applied.
All the experiments were performed under the
PyTorch framework using six RTX 4090 GPUs.

\begin{figure*}[hbt]
  \centering
  \includegraphics[width=1.0\linewidth]{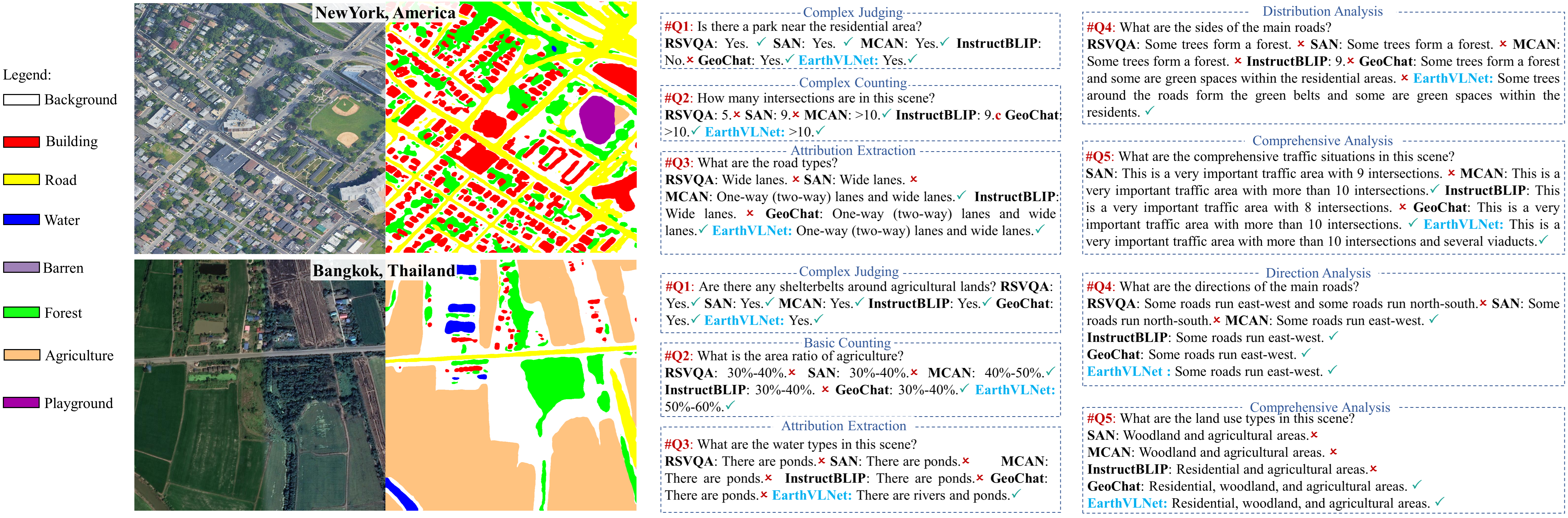}
  \vspace{-0.5cm}
  \caption{Comparative
    semantic segmentation and multiple-choice VQA results.
    The segmentation-guided methods performs better on complex questions. 
    EarthVLNet achieves better answer consistency across tasks and mitigates the negative effects of segmentation faults, to a certain extent.
    } 
  \label{fig:vis-result}
  \vspace{-0.5cm}
\end{figure*}

\noindent \textbf{Comparative Results.}
Thanks to the diverse questions,
\textit{EarthVLSet} can measure multiple perspectives of VQA models.
Tab~\ref{tab:vqa_comp_res} shows that all the methods perform well on the
basic questions, but show a lower performance on the complex questions.
The zero-shot evaluation results of GPT-4o and Claude achieve low accuracy  due to the domain gap between general requirements and remote sensing applications.
The models using pixel-level
segmentation features consistently obtain higher performances, especially for the
counting tasks. 
This is because the semantic locations provide more spatial details, which benefits the object statistics.
Due to the task similarity, the instruction-tuned models (InstructBLIP and GeoChat) outperform the models pre-trained by only causal language modeling tasks.
Because GeoChat was fine-tuned on large-scale remote sensing vision-language datasets, it has more transferability on the \textit{EarthVLSet}.
Equipped with similar or lower complexity, SOBA significantly outperforms 
the reference methods, especially for the relational reasoning questions.
Compared to SOBA, \textit{EarthVLNet} consistently
improves the performances of most sub-tasks, and the counting errors are further reduced.
Without semantic guidance, \textit{EarthVLNet} still achieves competitive results, due to our 
tailored optimization, especially for the counting tasks.
Guided by pixel-level semantic features,
the large multi-modal model can also show promising results in a conditional generation way.

\begin{figure}[hbt]
  \centering
  \includegraphics[width=1\linewidth]{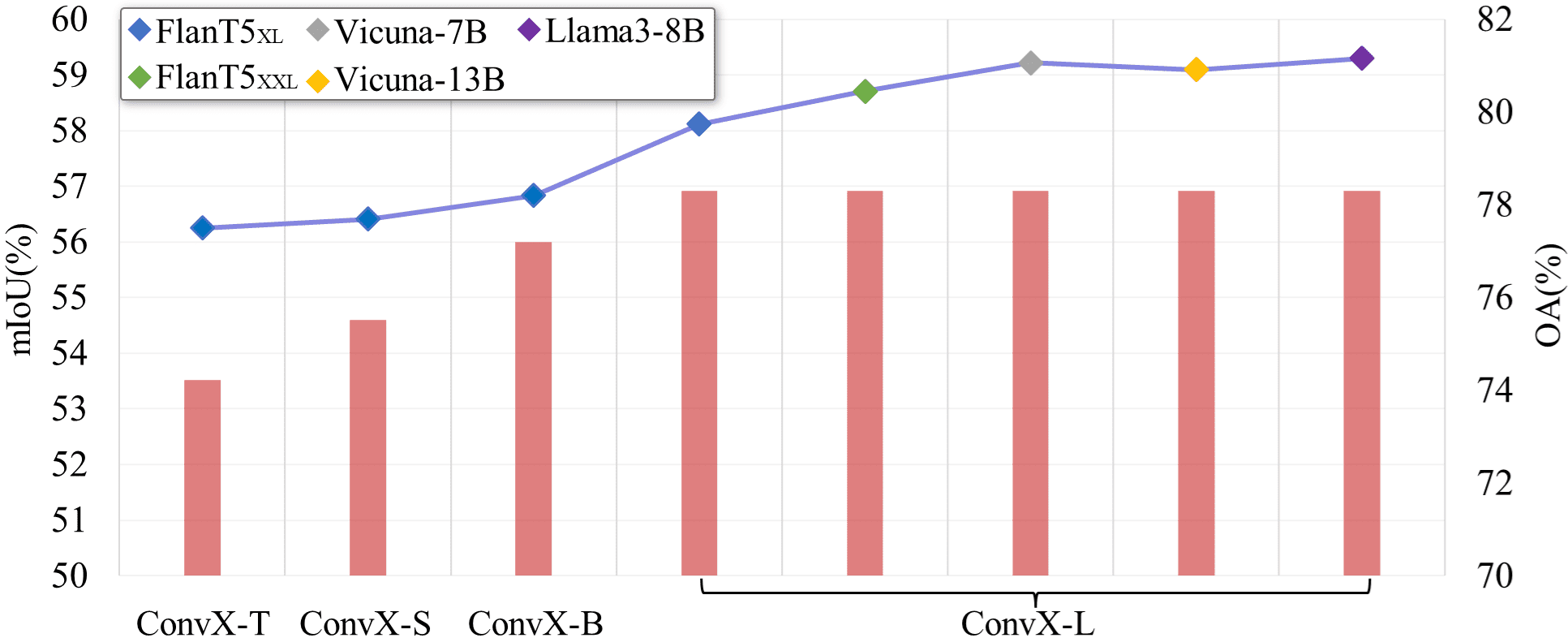}
  \vspace{-0.4cm}
  \caption{Ablation study of the vision and language modules for the multiple-choice task. 
  The VQA performance benefits from powerful vision encoders but exhibits less sensitivity to LLMs
  } 
  \label{fig:mc-encdec}
  \vspace{-0.4cm}
\end{figure}

\noindent \textbf{Comparative Visualizations.}
From the qualitative results shown in Fig.~\ref{fig:vis-result},
we found that most methods achieve the correct answers for
the relatively easy judging questions.
Due to the pixel-wise guidance,
the segmentation-guided methods achieve better performances on
the counting and more difficult questions.
As for the New York sample,
\textit{EarthVLNet} shows better semantic
consistency between the complex counting and comprehensive analysis questions.
Specifically,
the comprehensive analysis reveals that the reference methods fail to recognize the viaducts located in the top-right corner, whereas \textit{EarthVLNet} successfully identifies them.
Regarding the Bangkok sample, 
some parts of the agriculture class are misclassified into the road class 
so that the reference misjudges the direction.
However, \textit{EarthVLNet} is not negatively affected by the segmentation, demonstrating its robustness.


\noindent \textbf{Scalable Vision and Language Modules.}
To evaluate the effects of different vision and language modules,
we scaled up each part separately, with the results shown in Fig.~\ref{fig:mc-encdec}.
It is evident that better visual features lead to a higher VQA performance, especially in the counting tasks. 
This is because more accurate semantics improve the object localization and categorization, directly benefiting the downstream VQA task. 
Moreover, the choice of LLM also influences the VQA performance. 
Language models with stronger reasoning abilities on general tasks consistently perform better on the \textit{EarthVLSet}. 
Notably, changes to the vision components have a greater impact than changes to the language components, 
highlighting the importance of the visual features provided by the segmentation network.


\begin{table}[!hbt]
  \vspace{-0.3cm}
  \centering
  \caption{Comparative results with different fusing attentions} \label{tab:attention}
  \begin{tabular}{l|l|cc}
  \hline
  Object Guidance & Attention Type & $\uparrow$ OA (\%) & $\downarrow$ OR \\ \hline
  Only features &   -   & 79.97 & 2.582 \\ \hline
  Concat    & Spatial    & 80.21 & 2.543 \\
  +SA\cite{woo2018cbam}    & Spatial  & 79.83 & 2.590  \\
  +SCSE\cite{roy2018recalibrating}     & Channel\&Spatial & 80.37 & 2.551  \\
  +CBAM\cite{woo2018cbam}     & Channel\&Spatial  & 80.44 & 2.536  \\
  +SE\cite{hu2018squeeze}     & Channel & 80.72 & 2.439  \\
  +GC\cite{cao2019gcnet}     & Channel & 80.63 & 2.463  \\
  +OGA (ours)   & Channel     & 81.06 & 2.340  \\ \hline
  \end{tabular}
\vspace{-0.1cm}
\end{table}

\noindent \textbf{Object Guided Attention.}
The OGA effectively aligns the intermediate semantic features and pseudo masks into the same latent space. 
The existing attention mechanisms can be divided into three types according to the feature dimension.
Tab.~\ref{tab:attention} lists the results for the spatial, channel,
and hybrid attentions.
Compared to the spatial attention mechanisms, the channel attention mechanisms achieve more consistent improvements.
The dimensional concatenation of pseudo masks and visual features poses a challenge for 
spatial attention, which makes it difficult to calibrate the subspaces of visual features and object masks.
In contrast, channel attention enhances 
the key object semantics and diminishes the prominence of irrelevant features. 
Consequently, 
the OGA discards the spatial attention, resulting in superior accuracy.


\begin{table}[!hbt]
  \vspace{-0.2cm}
  \centering
  \caption{Comparative results with different optimization strategies} \label{tab:loss}
  \begin{tabular}{l|cc}
  \hline
  Optimization & $\uparrow$ OA (\%) & $\downarrow$ OR \\ \hline
  CE loss  & 79.91 & 2.591 \\ \hline
  Focal loss\cite{lin2017focal}   & 80.24 & 2.527  \\
  DIW loss\cite{rajpurkar2017chexnet}    & 79.51 & 2.654 \\
  OHEM\cite{shrivastava2016training}    & 80.44 & 2.481  \\
  SOM\cite{FactSeg}     & 80.19 & 2.536  \\ \hline
  ND-Shared   & 80.63 & 2.422  \\ 
  ND-Separated (ours) & 81.06 & 2.340  \\   \hline  
\end{tabular}
\vspace{-0.1cm}
\end{table}


\noindent \textbf{Comparative Results on Other Datasets.}
To evaluate the model generalizability,
we conducted more comparative experiments on other VQA datasets (Tab.~\ref{tab:comp_other}).
As there are no matched semantic masks for the RSVQA dataset,
the semantic features were generated by Semantic-FPN (ConvX-L) trained on \textit{EarthVLSet}.
\textit{EarthVLNet} outperforms all the reference methods and
shows strong generalizability on the different datasets.
It is notable to find that, even in cross-dataset scenarios,
the segmentation features also demonstrate their significance for VQA guidance.

\begin{table}[!hbt]
  \centering
  \caption{Comparative results on other VQA datasets} \label{tab:comp_other}
  \resizebox{1.0\linewidth}{!}{
  \begin{tabular}{l|c|c|ccc}
  \hline
  \multirow{2}{*}{Method} & \multirow{2}{*}{Seg} & \multirow{2}{*}{Params} & \multicolumn{3}{c}{$\uparrow$ OA (\%)}  \\ \cline{4-6}
        &        &         & FloodNet\cite{rahnemoonfar2021floodnet}  & EarthVQA\cite{wang2024earthvqa} & RSVQA\cite{lobry2020rsvqa} \\ \hline
  \multicolumn{2}{l|}{\rsb \textit{Classification-based}} &&& \\
  MAC\cite{hudson2018compositional}     & $\times$       & 49.9M  & 79.86   & 73.49  & 84.11     \\
  RSVQA\cite{lobry2020rsvqa}     & $\times$    & 34.9M        & 80.13   & 73.67  & 83.19     \\
  RSIVQA\cite{zheng2021mutual}    & $\times$   & 72.5M        & 79.52   & 76.44  & 77.90     \\
  SAN\cite{yang2016stacked}     & $\times$   & 37.3M        & 79.78    & 76.50 & 83.96     \\
  BAN\cite{kim2018bilinear}     & $\checkmark$   & 30.2M        & 79.84   & 77.23  & 85.15     \\
  MCAN\cite{yu2019mcan}      & $\checkmark$    & 17.7M        & 80.74   & 78.38  & 85.29     \\
  BUTD\cite{Anderson_2018_CVPR}      & $\checkmark$  & 12.3M        & 81.14  & 78.25 & 85.59     \\
  LXMERT\cite{tan2019lxmert}    & $\checkmark$   & 87.6M        & 80.69    & 77.94 & 85.44     \\ 
  D-VQA\cite{wen2021debiased}     & $\checkmark$   & 17.6M        & 79.98  & 77.80 & 84.21     \\ \hline
  \multicolumn{2}{l|}{\gb \textit{Generation-based}} &&& \\
  ALBEF\cite{li2021align}     & $\times$     & 290.6M        & 80.41   & 74.80  & 83.57   \\ 
  BLIP-2\cite{BLIP2}    & $\times$       & 3.9B        & 79.33   & 74.07  & 82.94     \\ 
  InstructBLIP\cite{liu2023improved}  & $\times$   & 4.0B        & 81.22   & 76.25  & 84.80     \\ 
  LLaVANeXT\cite{liu2024llavanext}   & $\times$    & 7.2B        & 81.89   & 78.17  & 85.25   \\ 
  GoeChat\cite{kuckreja2023geochat}     & $\times$   & 7.2B        & 81.37   & 77.91  & 85.28   \\ \hline
  \rsb \textcolor{revision}{SOBA~\cite{wang2024earthvqa}}   & $\checkmark$    & 19.9M    & \underline{82.77}    & \underline{78.49} & \underline{85.81}     \\ 
  \gb EarthVLNet  & $\checkmark$      & 6.9B   & \textbf{83.84}   & \textbf{79.26}  & \textbf{86.21}   \\ 
  \hline
  \end{tabular}}
  \end{table}

\vspace{-0.2cm}
\subsection{Open-Ended Visual Question Answering}
\noindent\textbf{Implementation Details.}
As most of the small-scale VQA methods are unable to generate open-ended answers, we focused on large vision-language generative models in the experiments.
All the open-ended VQA models (except for GPT4-o and Claude) were trained for
20k steps with a batch size of 16.
After trial experiments, the initial learning rate was set to 1e-5, and the other settings remained the same as in the multiple-choice implementations.
The traditional \cite{touvron2023llama}, LLM-based\cite{manas2024improving}, human-based\cite{elangovan2024considers} metrics were adopted for 
reporting the performances.
For human-based evaluation, we hired 10 urban planning experts to rate 20,000 samples (21.86\% of the Test set) on accuracy, relevance, and completeness using a 5-point scale. 
The detailed rating criteria is provided in the Appendix B.
As we had five synonymous ground truths,
the mean metrics were calculated based on all the labeled answers.

\begin{table*}[hbt]
  \caption{Open-ended VQA benchmarks on general-purposed and remote sensing-tailored VLMs} \label{tab:open-ended}
  \centering
  \resizebox{1.0\linewidth}{!}{
  \begin{tabular}{l|c|c|ccccccccc}
  \hline
  Method  & Seg & Params & $\uparrow$BLEU1   & $\uparrow$BLEU2   & $\uparrow$BLEU3   & $\uparrow$BLEU4   & $\uparrow$METEOR & $\uparrow$ROUGE-L  & $\uparrow$CIDEr  & $\uparrow$LAVE\tiny{FT}(\%) & $\uparrow$Human\\ \hline
  BUTD\cite{Anderson_2018_CVPR} & $\checkmark$  & 43.6M   & 0.5124 & 0.3667 & 0.2718 & 0.2062 & 0.2511 & 0.3790  & 0.2788 &76.74 & 3.66\\
  LXMERT\cite{tan2019lxmert}  & $\checkmark$ & 114.3M   & 0.5393 & 0.3878 & 0.2869 & 0.2156 & 0.2429 & 0.3869 & 0.3031 & 76.42 & 3.73\\
  ALBEF\cite{li2021align}    & $\times$ & 290.6M   & 0.5058 & 0.3515 & 0.2495 & 0.1797 & 0.2365 & 0.3741 & 0.2564 & 67.21 & 3.11 \\
  BLIP-2\cite{BLIP2}   & $\times$ & 3.9B   & 0.4777 & 0.3298 & 0.2344 & 0.1684 & 0.1871 & 0.3419 & 0.2015 & 65.43 & 3.17\\
  InstructBLIP\cite{liu2023improved} &$\times$ & 4.0B   & 0.5491 & 0.3989 & 0.2965 & 0.2204 & 0.2276 & 0.3938 & 0.3392 & 70.69 & 3.35\\
  GeoChat\cite{kuckreja2023geochat} &$\times$ & 7.2B   & 0.5610 & 0.4108 & \underline{0.3118} & 0.2373 & 0.2489 & 0.3925 & 0.3504 & 73.61& 3.44\\
  ViP-LLaVA\cite{cai2024vip} &$\times$ & 7.2B   & 0.5601 & 0.4094 & 0.3010 & 0.2294 & 0.2316 & 0.3958 & 0.3492 &72.80 & 3.64\\
  LLaVANeXT\cite{liu2024llavanext}  & $\times$& 7.2B  & 0.5619 & 0.4128 & 0.3106 & 0.2366 & 0.2493 & \underline{0.3994} & 0.3520 & 72.69 & 3.17\\ \hline
  GPT-4o\cite{liu2024llavanext}  & $\times$& -  & 0.2111 & 0.1327 & 0.0184 & 0.0127 & 0.1392 & 0.1871 & 0.1239 & 56.74& 2.81\\ 
  Claude3 Opus\cite{liu2024llavanext}  & $\times$& -  & 0.2564 & 0.1532 & 0.0257 & 0.0229 & 0.1270 & 0.2551 & 0.1844 &  59.22 & 2.94\\ \hline
  EarthVLNet w.o. seg &$\times$ & 6.9B & \underline{0.5653} & \underline{0.4140} & 0.3115 & \underline{0.2417} & \underline{0.2496} & 0.3976 & \underline{0.3552} & \underline{77.94} & 3.98 \\ 
  EarthVLNet  &$\checkmark$ & 6.9B   & \textbf{0.5726} & \textbf{0.4229} & \textbf{0.3211} & \textbf{0.2483} & \textbf{0.2520} & \textbf{0.4025} & \textbf{0.3661}  & \textbf{80.44} & \textbf{4.25}\\ 
  \hline
  \end{tabular}}
\end{table*}
\begin{figure*}[hbt]
  \centering
  \includegraphics[width=1.0\linewidth]{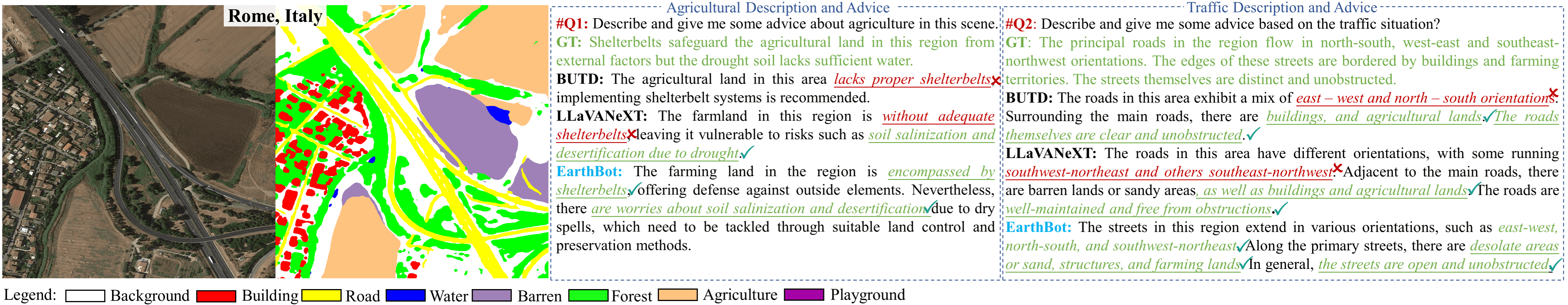}
  \vspace{-0.5cm}
  \caption{Comparative semantic segmentation and open-ended VQA results. The LLMs exhibit superior performances in long sentence generation, compared to the small-scale generative models, 
  and are more suitable for the open-ended VQA task with HSR imagery. 
  }
  \vspace{-0.5cm} 
  \label{fig:vis-openresult}
\end{figure*}

\noindent\textbf{Comparative Results.}
Tab.~\ref{tab:open-ended} presents a comparative analysis of the advanced multi-modal generation methods. 
In the context of comparable model parameters, LEXMERT demonstrates a superior performance, 
exceeding ALBEF by 0.03, as measured by BLEU1.
Furthermore, the proposed \textit{EarthVLNet} demonstrates a performance enhancement of 0.05 
when compared to its counterpart without segmentation features.
These two cases substantiate the importance of objectness semantics, 
aligning with the findings observed in the multiple-choice evaluations.
Compared with the small-scale models, the large VLMs 
exhibit a markedly superior performance in open-ended VQA tasks.
With regard to the generation of long answers, 
the abilities of LLMs for induction and conclusion are critical.


\noindent \textbf{Comparative Visualizations.}
Fig.~\ref{fig:vis-openresult} provides a visual comparison of the open-ended VQA task predictions via a representative test sample from Rome in Italy.
Regarding ``Agricultural Description and Advice'', the ground truth emphasizes two critical elements: shelterbelts and arid agricultural land. 
BUTD, as a typical small-scale method, fails to identify the shelterbelts and misses the agricultural context. 
LLaVANeXT correctly addresses the soil drought concerns but overlooks the presence of shelterbelts. 
The proposed \textit{EarthVLNet} accurately describes the situational elements and provides reasonable advice.
In the ``Traffic Description and Advice'' task, the ground truth delineates road orientations, nearby objects, and traversability status. 
Both BUTD and LLaVANeXT successfully describe the passable roads and adjacent buildings as well as the agricultural lands, but misinterpret the road orientations. 
Conversely, the proposed \textit{EarthVLNet} demonstrates a superior accuracy in its responses.
\textit{EarthVLNet} achieves the best performances on all traditional, LLM-based, and Human-based metrics, 
demonstrating its superiority comprehensively.

\begin{figure}[hbt]
  \centering
  \includegraphics[width=1\linewidth]{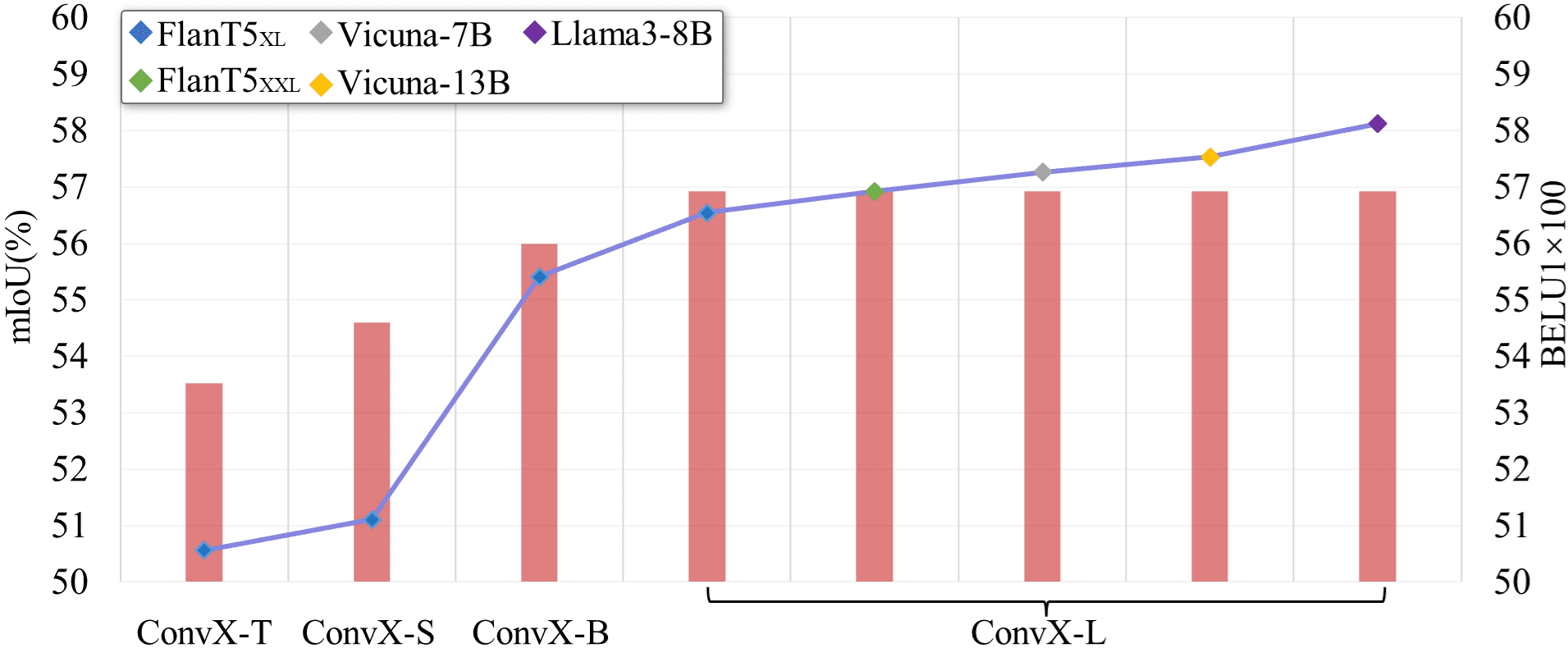}
  \vspace{-0.4cm} 
  \caption{Ablation study of vision and language modules for the open-ended task.
  The optimal 
  performance necessitates both robust vision encoders and advanced language decoders.
  }
  \vspace{-0.4cm} 
  \label{fig:od-encdec1}
\end{figure}

\noindent \textbf{Scalable Vision and Language Modules.} As shown in Fig.~\ref{fig:od-encdec1},
in contrast to the multiple-choice tasks where visual encoders predominantly influence the overall performance, the open-ended tasks demonstrate sensitivity to both the vision and language modules.
When the language decoder is fixed as FlanT5{\tiny XL}, scaling up the ConvNeXt encoders results in a BLEU1 increase from 0.505 to 0.564. Furthermore, with the segmentation results fixed at mIoU=56.92\% using ConvNeXt-Large, 
the integration of more sophisticated LLMs yields additional improvements in overall VQA performance.
These findings underscore the critical importance of both vision and language modules in optimizing open-ended VQA performance.

\section{Application of Urban Heat Island}

\begin{figure}[hbt]
  \vspace{-0.2cm} 
  \centering
  \includegraphics[width=1\linewidth]{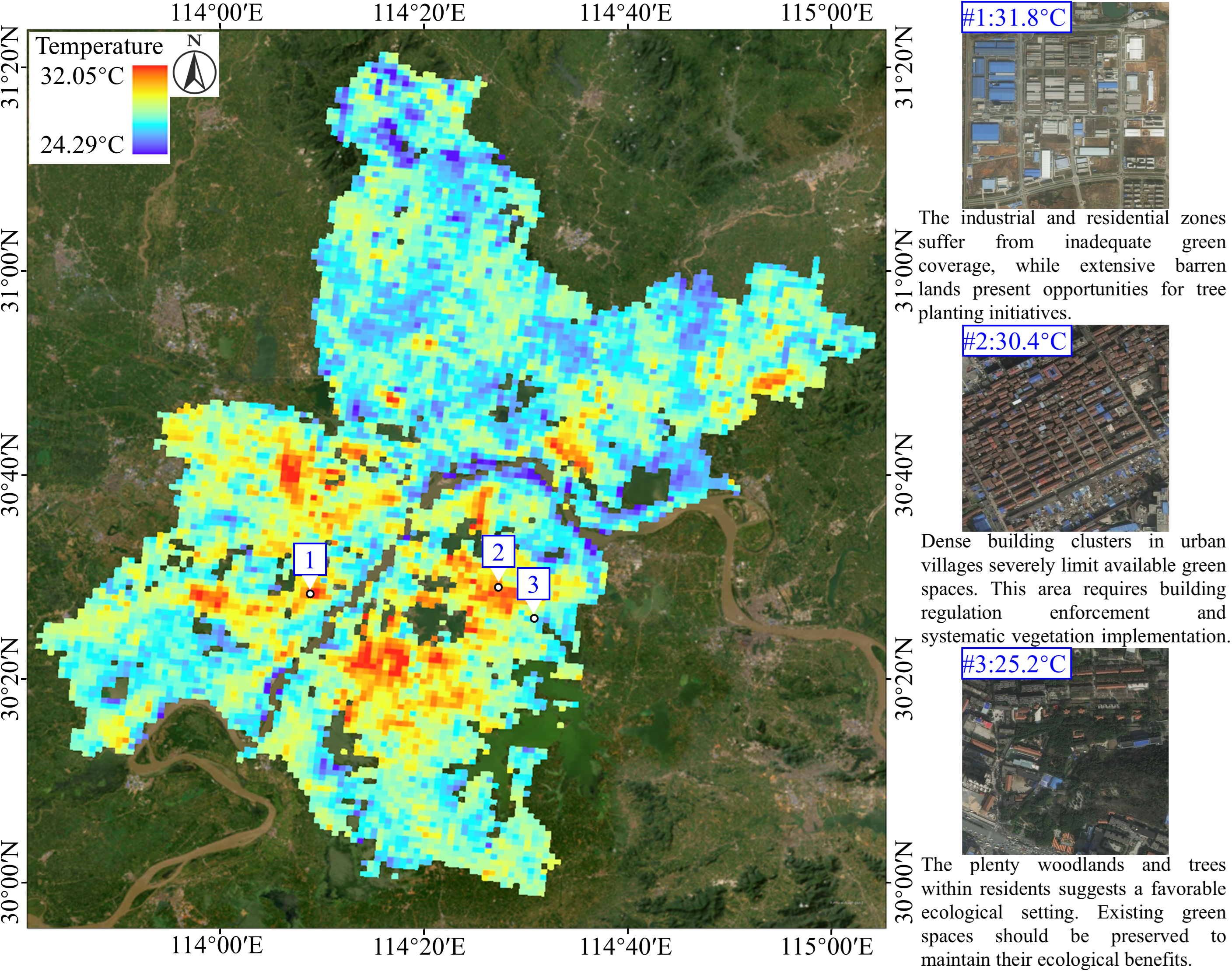}
  \vspace{-0.6cm} 
  \caption{
  Application of urban heat island in Wuhan City, China.
  By answering the open-ended question `Describe and give me some advice about the greening renovation.', the chosen three samples 
  show the guiding significance of alleviating heat island effects.
  }
  \vspace{-0.6cm} 
  \label{fig:heat_island}
\end{figure}

This section discusses the applicability of \textit{EarthVLNet} via 
urban heat island effects.
Urban green spaces mitigate heat exposure risks, yet their distribution remains imbalanced amid rapid urbanization\cite{yin2023unequal}.
Combined with the monthly temperature product\cite{zhang2023hitic},
we utilized \textit{EarthVLNet} to obtain the greening renovation advice.
Fig.~\ref{fig:heat_island} illustrates the mean apparent temperature distribution across Wuhan in July 2020.
Besides, three diverse samples are selected to show \textit{EarthVLNet}'s responses to the question `Describe and give me some advice about the greening renovation.'
Region \#1 denotes the industrial area with high temperatures resulting from intensive machinery operations and industrial emissions.
Strategic tree planting in bare land can significantly mitigate the thermal stress experienced by nearby residents.
Region \#2 represents an urban village with dense low-rise buildings, preventing air circulation.
It is reasonable that \textit{EarthVLNet} proposes building regulation enforcement and systematic vegetation implementation.
In contrast, Region \#3 showcases a favorable ecological environment that should be protected.
According to \textit{EarthVLNet}'s analysis, 612 communities in Wuhan City require green space enhancement, with 81\% of these areas exhibiting severe urban heat island effects (temperatures exceeding $30^\circ$C).
Based on these results, city planners could identify critical areas requiring attention quickly.

In this case, \textit{EarthVLNet} support micro- and macro-level green space analysis efficiently, providing reasonable advices to 
mitigate heat exposure risks in megacities.

\section{Conclusion}

In this paper, we present \textit{EarthVLSet}, a multi-task vision-language dataset containing 734k co-paired "image-mask-QA pairs," 
and \textit{EarthVLNet}, a large vision-language model that progressively integrates semantic segmentation and VQA capabilities. 
Our framework combines pixel-wise semantic understanding with LLM-powered relational reasoning,
 enhanced by object counting optimization for remote sensing scenes. 
 Comprehensive evaluations demonstrate \textit{EarthVLNet}'s effectiveness in Earth vision understanding while identifying 
 three directions for future development. This work establishes a robust foundation for advancing geographical applications in the Earth vision field.
\vspace{-0.3cm}
\printglossary[type=\acronymtype]

{\small
\bibliographystyle{IEEEtran}
\bibliography{main}
}



\begin{IEEEbiography}[{\includegraphics[width=1in,height=1.25in,clip,keepaspectratio]{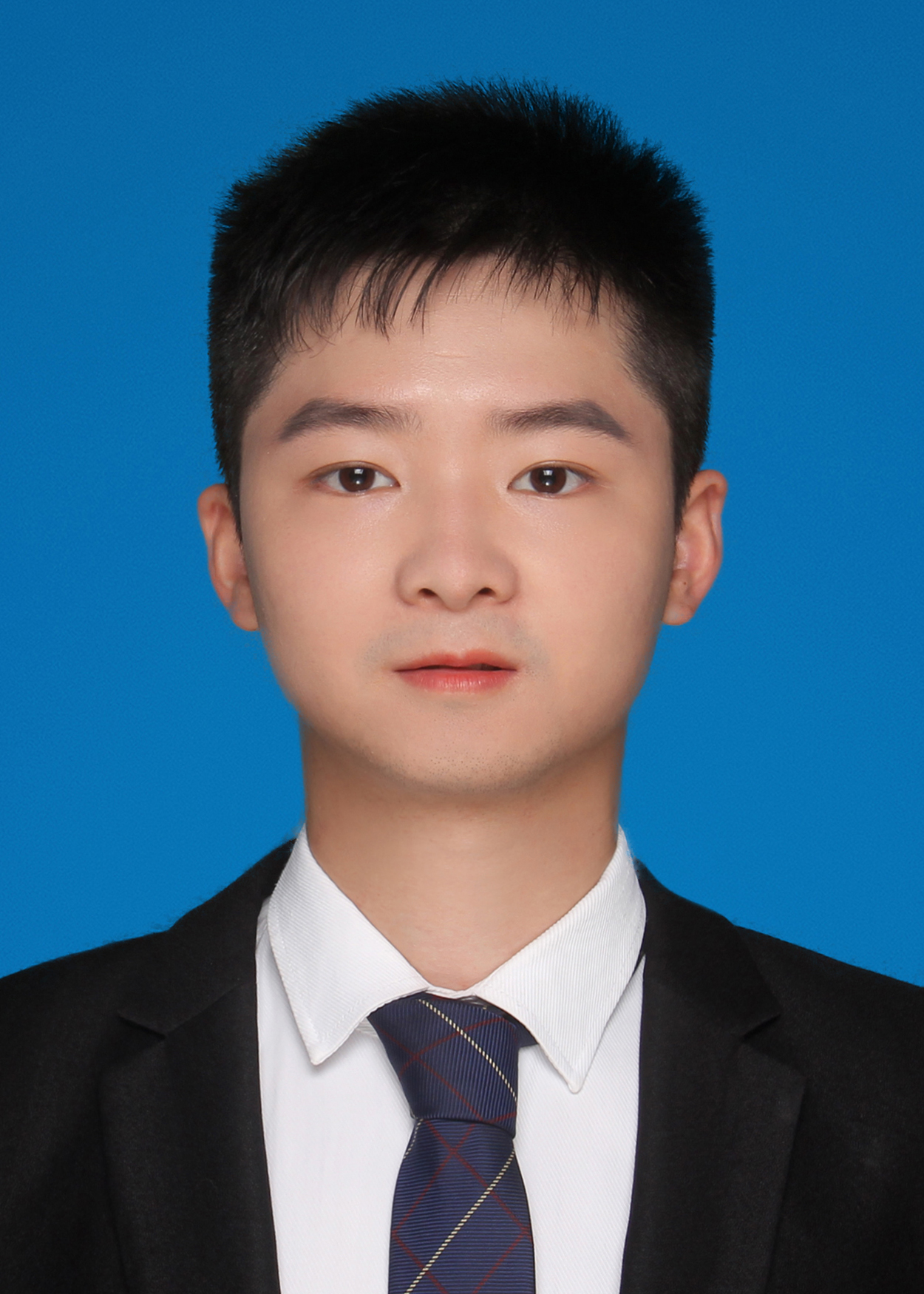}}]{Junjue Wang}
received the B.S. degree from the School of Geography and Information Engineering, China University of Geosciences, Wuhan, China, in 2019 and the doctoral degree in photogrammetry and remote sensing from Wuhan University, Wuhan, in 2024. 
He is currently a project researcher at the Department of Complexity Science and Engineering, The University of Tokyo.
His major research interests are multi-modal remote sensing data processing. 

He won 1st prize in 2025 AI for Earthquake Response Challenge, 1st in 2022 LandSlide4Sense Contest, the 2rd prize in Single-view Semantic 3D Challenge of 2019 IEEE GRSS Data Fusion Contest, the 4th in xView2 Challenge.
\end{IEEEbiography}
\begin{IEEEbiography}[{\includegraphics[width=1in,height=1.25in,clip,keepaspectratio]{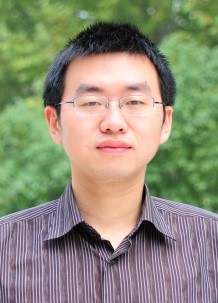}}]{Yanfei Zhong}
received the B.S. degree in information engineering and the Ph.D. degree in photogrammetry and remote sensing from WuhanUniversity, China, in 2002 and 2007, respectively. Since 2010, He has been a Full professor with the State Key Laboratory of Information Engineering in Surveying, Mapping and Remote Sensing (LIESMARS), Wuhan University, China. He organized the Intelligent Data Extraction, Analysis and Applications of Remote Sensing (RSIDEA) research group. 

He is a Fellow of the Institution of Engineering and Technology (IET). He was a recipient of the 2016 Best Paper Theoretical Innovation Award from the International Society for Optics and Photonics. He won the Second-Place Prize in 2013 IEEE GRSS Data Fusion Contest and the Single-view Semantic 3-D Challenge of the 2019 IEEE GRSS Data Fusion Contest, respectively. He is currently serving as an Associate Editor for the IEEE Journal of Selected Topics in Applied Earth Observations abd Remote Sensing, and the International Journal of Remote Sensing.
\end{IEEEbiography}
\begin{IEEEbiography}[{\includegraphics[width=1in,height=1.25in,clip,keepaspectratio]{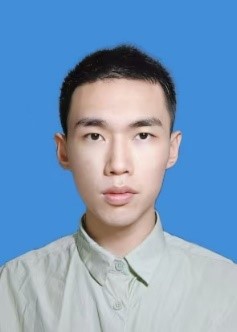}}]{Zihang Chen}
received the B.S. degree from the Wuhan University, Wuhan, China, in 2023. He is currently pursuing a Master's degree in Photogrammetry and Remote Sensing at Wuhan University. His major research interests are multi-modal remote sensing data interpretation.
He was selected as the finalist of the student paper competition at the 2022 IEEE International Geoscience and Remote Sensing Symposium (IGARSS).
\end{IEEEbiography}
\begin{IEEEbiography}[{\includegraphics[width=1in,height=1.25in,clip,keepaspectratio]{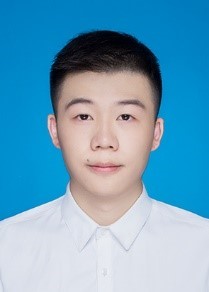}}]{Zhuo Zheng}
received the BS degree from the School of Geography and Information Engineering, China University of Geosciences, Wuhan, China, in 2018 and the PhD degree in photogrammetry and remote sensing from Wuhan University, Wuhan, in 2023. 
He is currently a postdoctoral researcher at the Stanford Artificial Intelligence Laboratory (SAIL), Department of Computer Science, Stanford University. His major research interests Earth vision and simulation, especially multi-modal, and multitemporal remote sensing image analysis. 
He has published over 10 first-author papers in leading journals and conferences, such as IEEE TPAMI, IJCV, RSE, ISPRS P\&RS, NeurIPS, IEEE CVPR, ICCV, IEEE TGRS, etc.

He won the second place prize in the Single-view Semantic 3D Challenge of the 2019 IEEE GRSS Data Fusion Contest, the fourth place overall in xView2 Challenge, the top graduate award in SpaceNet 6 and EarthVision workshop challenge at CVPR 2020, the fourth place in Multitemporal Semantic Change Detection Challenge of 2021 IEEE GRSS Data Fusion Contest, and the fifth place and Model Write-Up Bonus Award in Overhead Geopose Challenge. He is also first-place recipient of the 2021 John I. Davidson President's Award.
\end{IEEEbiography}
\begin{IEEEbiography}[{\includegraphics[width=1in,height=1.25in,clip,keepaspectratio]{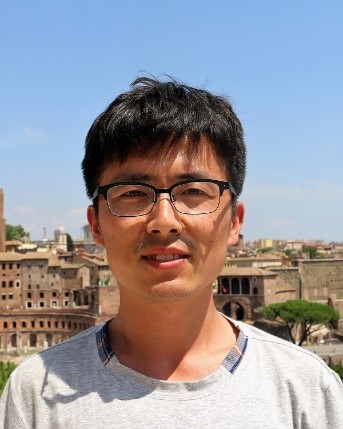}}]{Ailong Ma}
received the B.S. degree from the China University of Petroleum, Qingdao, China, in 2010, and the Ph.D. degree in photogrammetry and remote sensing from the Wuhan University, Wuhan, China, in 2017. He is currently working as a Research Associate with Wuhan University. His major research interests are remote sensing image processing, evolutionary computing, and pattern recognition.
\end{IEEEbiography}

\begin{IEEEbiography}[{\includegraphics[width=1in,height=1.25in,clip,keepaspectratio]{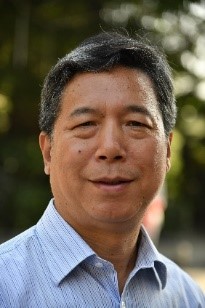}}]{Liangpei Zhang}
received the B.S. degree in physics from Hunan Normal University, Changsha, China, in 1982, the M.S. degree in optics from the Xi'an Institute of Optics and Precision Mechanics, Chinese Academy of Sciences, Xi'an, China, in 1988, and the Ph.D. degree in photogrammetry and remote sensing from Wuhan University, Wuhan, China, in 1998. He is a “Chang-Jiang Scholar” chair professor appointed by the ministry of education of China in LIESMARS, Wuhan University. He published more than 700 research papers and five books. He is the Institute for Scientific Information (ISI) highly cited author. His research interests include hyperspectral remote sensing, high-resolution remote sensing, image processing, and artificial intelligence.

Dr. Zhang is a Fellow of Institute of Electrical and Electronic Engineers (IEEE) and the Institution of Engineering and Technology (IET). He was a recipient of the 2010 best paper Boeing award, the 2013 best paper ERDAS award from the American society of photogrammetry and remote sensing (ASPRS) and 2016 best paper theoretical innovation award from the international society for optics and photonics (SPIE). His research teams won the top three prizes of the IEEE GRSS 2014 Data Fusion Contest. 
He is currently serving as an associate editor of the IEEE Transactions on Geoscience and Remote Sensing.
\end{IEEEbiography}







\end{document}